\documentclass[twocolumn]{galbot}

\usepackage{amsmath,amssymb}
\usepackage{array}
\usepackage{adjustbox}
\usepackage{booktabs}
\usepackage{flafter}
\usepackage{placeins}
\usepackage{makecell}
\usepackage{pifont}
\usepackage{colortbl}
\usepackage{xspace}

\makeatletter
\DeclareRobustCommand\onedot{\futurelet\@let@token\@onedot}
\def\@onedot{\ifx\@let@token.\else.\null\fi\xspace}

\makeatother

\renewcommand{\paragraph}[1]{\vspace{1.25mm}\noindent\textbf{#1}}

\hypersetup{
  colorlinks=true,
  linkcolor=galbotbg,
  citecolor=galbotbg,
  urlcolor=galbotbg
}

\title{RoboGesture: Real-Time Semantic-aligned Co-Speech Gestures Generation for Humanoid Interaction}

\author[1,2,*]{Zifan Wang}
\author[2,3,*]{Ziang Ren}
\author[1,2,*]{Pengyang Shi}
\author[4]{Zirui Wang}
\author[2]{Chenghuai Lin}
\author[2]{Tianze Wang}
\author[1,2]{Zekun Qi}
\author[4]{Liangliang Zhao}
\author[2,5]{He Wang}
\author[1,2,6,\dagger]{Li Yi}

\affiliation[1]{Tsinghua University}
\affiliation[2]{Galbot Inc.}
\affiliation[3]{Beijing Institute of Technology}
\affiliation[4]{Harbin Institute of Technology}
\affiliation[5]{Peking University}
\affiliation[6]{Shanghai Qi Zhi Institute}

\contribution[*]{Equal contribution}
\contribution[\dagger]{Corresponding author}

\date{\today}
\page{\url{https://RoboGesture.github.io}}

\abstract{
Enabling humanoid robots to respond to human speech with synchronized and semantically meaningful gestures is fundamental to natural human-robot interaction. However, this task faces three critical barriers: the scarcity of semantically rich datasets, the ``modality eclipse'' where models ignore audio cues in favor of kinematic inertia, and the sim-to-real gap regarding physical safety. We propose RoboGesture, a robot-centric framework that co-designs data, modeling, and control to power a complete interactive human--humanoid system in which the robot listens, responds, and gestures in real time.
We first establish the RoboGesture dataset featuring over 300 gesture categories and develop an automated pipeline to synthesize large-scale collision-free, robot-specific audio--motion pairs. Our architecture features a Hierarchical Semantic-Acoustic Aligner that extracts multi-granular prosodic and semantic cues directly from raw audio tokens. These cues drive a Streaming Conditional Motion Generator based on a diffusion transformer with conditional flow matching. To ensure high responsiveness, we introduce Anti-Inertia CFG Masking, which prevents the model from collapsing into repetitive historical patterns by compelling it to proactively mine control signals from the audio modality. Finally, an MPC-based safety filter ensures real-time, collision-free execution on physical hardware. Experiments on a Unitree G1 humanoid demonstrate that RoboGesture generates safer, more rhythmic, and more semantically appropriate responses compared to state-of-the-art baselines.
}

\renewcommand{\posttitle}{%
  \vspace{2pt}
  \begingroup
    \centering
    \captionsetup{type=figure}
    \includegraphics[width=0.96\textwidth]{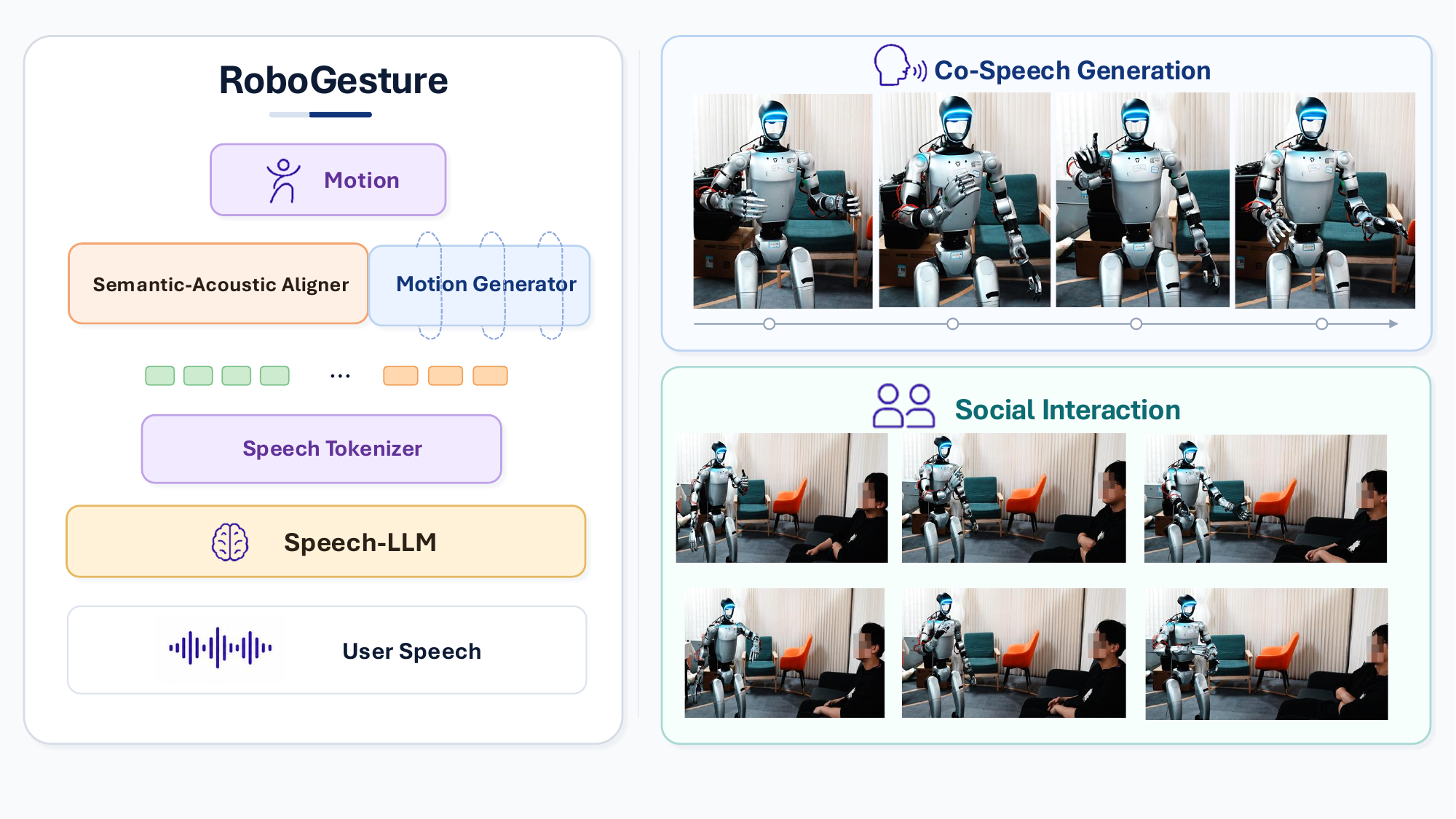}\par
    \vspace{-2pt}
    \captionof{figure}{RoboGesture generates expressive, semantically aligned, and
    safety-aware humanoid co-speech gestures from streaming speech for real-time
    social interaction.}
    \label{fig:eccv_teaser}
    \vspace{-2pt}
  \endgroup
}

\begin{document}
\raggedbottom

\maketitle
\pagestyle{empty}
\clearpage

\section{Introduction}
\label{sec:intro}

The ability of a humanoid robot to \emph{listen} to human speech and \emph{respond} with synchronized, semantically meaningful audio and body motion lies at the heart of natural human-robot interaction (HRI). Such capability elevates robots from mere tools to genuine partners in assistive care, education, and collaboration. This work pursues a complete system that endows a humanoid with expressive, embodiment-safe, and real-time audio-motion responses—making face-to-face conversation with machines not only possible but intuitive and engaging.

Realizing this vision, however, confronts three entrenched barriers. First, the scarcity of large-scale, semantically rich audio-motion datasets with fine-grained annotations~\cite{liu2022beat,ghorbani2023zeroeggs} limits models' capacity to establish semantic alignment between speech and gestures. Existing methods therefore resort to post-processing heuristics or text-based intermediate representations~\cite{zhang2025semtalk,zhang2024semantic,Zhi_2023_ICCV}, yet these compromises inherently sacrifice semantic and temporal coherence in the generated motion. Second, these limitations become critically exacerbated in online generation settings: without explicit modeling of audio semantics and controllable generation mechanisms, models tend to collapse into motion shortcuts—excessively replicating historical motion patterns rather than responding to incoming acoustic cues. Third, the sim-to-real gap between virtual avatars and physical humanoids remains substantial. Directly retargeting generated human motions to humanoid robots via online retargeting and PD control often yields unstable, unsafe behaviors, including jerky trajectories and self-collisions.

In this work, we aim to design a robot-centric, data-model co-designed framework by tackling the above challenges. As shown in Figure~\ref{fig:eccv_teaser}, we present a comprehensive solution that enables \textbf{real-time}, \textbf{semantically aligned}, and \textbf{safety-aware} humanoid co-speech gesture synthesis in an \textbf{end-to-end manner}, and serves as the motion core of a complete interactive human–humanoid system in which the robot listens, responds, and gestures.

To address the scarcity of large-scale, tightly aligned audio-motion data, we build a large-scale, semantically rich dataset by capturing a library of expressive body and hand gestures guided by an established gesture taxonomy. We then develop an automatic synthesis pipeline to generate vast quantities of paired audio and motion sequences. Critically, this pipeline incorporates an offline motion retargeting step to map all human motions to our target humanoid's kinematics, followed by an MPC-based filter that vets the entire dataset to remove any potential self-collisions. This process yields a clean, extensive, and inherently safe training corpus of robot-specific motions.

To better capture temporal alignment between raw audio and semantics, we develop a hierarchical semantic-acoustic aligner.
Unlike text-based methods that lose vital prosodic cues like intonation and emphasis, our aligner processes raw audio through a multi-level transformer architecture. By decoupling audio into streaming low-level rhythmic pulses and high-level semantic labels, the aligner is able to extract "pre-phonetic" anticipatory signals—such as the body's energy accumulation before a shout—ensuring the robot's movements are synchronized with both the beat and the intent of the speech.

To ensure these cues effectively drive the robot and prevent it from collapsing into history motion shortcut, we develop a streaming conditional motion generator. We observe that models often "cheat" by relying excessively on past kinematic inertia, effectively ignoring the weaker audio and semantic guidance. To break this dependency, we introduce a two-stage training strategy with anti-inertia masking, which randomly masks historical context to compel the diffusion-based motion generator to proactively mine control signals from the audio modality. By integrating FiLM for global semantic tone and Cross-Attention for local rhythmic alignment, our framework enables seamless, end-to-end streaming from raw audio to safe, high-fidelity robot trajectories.

We validate our framework through extensive quantitative metrics and qualitative user studies focusing on motion quality, synchrony, naturalness, and safety. Furthermore, we demonstrate the system's efficacy through real-world social interactions on a physical humanoid robot. Our approach consistently generates behaviors with superior semantic and rhythmic alignment compared to existing baselines.
Our contributions are:(1)~A robot-centric, data-model co-designed framework for an interactive human–humanoid system that enables end-to-end, streaming generation of real-time, semantically aligned, and safety-aware co-speech gestures, and that we deploy as a complete listen–respond–gesture loop on a physical humanoid. (2)~A robust data synthesis pipeline that produces a large-scale, semantically rich, and collision-free training corpus by mapping expressive human gestures to robot-specific kinematics. (3)~A continuous audio-driven motion policy to ensure tight temporal synchronization and prevent the model from collapsing into repetitive historical motion patterns.

\section{Related Works}
\label{sec:relatedworks}

\subsection{Human-Humanoid Social Interaction}
% avatar: solami; Digital life project: Autonomous 3d characters with social intelligence. 
% works on avatars; real-world problems: real-time retargeting loses gesture details & safety
% HRI: planned-based motion interaction; decoupled audio and motion module 

Social interactions between humans and humanoid robots have gained increasing attention~\cite{matheus2025long,dafarra2024icub3,galatolo2025simultaneous,valls2025robot,wang2025mobileh2r,wang2024genh2r,cao2024ai}, driven by advances in both robotic embodiment and social intelligence modeling. 
Many works~\cite{jiang2025solami,Cai_2024_CVPR,Ao2023GestureDiffuCLIP,ijcai2023p650} have addressed social interaction problems through virtual avatars. For instance, SOLAMI~\cite{jiang2025solami} enables immersive interaction with 3D autonomous characters and aspires to bridge toward robotic embodiments. However, a significant domain gap exists between avatar environments and real robots. From the embodiment perspective, physical humanoids face hardware, safety, and real-time constraints, whereas virtual-avatar environments do not suffer from these limitations.  A core challenge in this direction is retargeting human motions to humanoid robots: classical pipelines~\cite{yoon2019robots} first learn models in human representation and then retarget to the robot, or learns human-to-robot mappings directly~\cite{mascaro2024robot}. Despite progress in high-fidelity retargeting~\cite{joao2025gmr,yang2025omniretarget}, balancing real-time performance, fine-grained gesture detail, and safety remains difficult in social scenarios. Motivated by these limitations, we instead learn directly in the robot’s representation, addressing cross-embodiment alignment as a pre-processing rather than post-processing problem.

For social intelligence modeling on real-world robots, traditional systems rely on rule-based or policy-structured controllers, including behavior trees and hand-crafted preconditions~\cite{bettosi2024systematic,scherf2024learning,tagliamonte2024generalizable}. While interpretable, such pipelines struggle to capture the open-ended complexity of real social behavior, and often decouple modalities—such as speech and gesture that humans naturally coordinate. 
%Recent multimodal end-to-end models~\cite{jiang2025solami,chen2024diffsheg,jiang2024motiongpt} demonstrate stronger behavior modeling and generalization, benefiting from large human datasets, but they do not account for the unique embodiment gap. 
In contrast, we develop a multimodal end-to-end model trained directly in robot space, enabling natural, coherent social interaction on a real humanoid platform.

\subsection{Co-speech Gesture Generation}

\begin{table*}[t]
\centering
\scriptsize
\caption{Comparison with related methods.\textit{Streaming}: whether the model has a design for streaming support. \textit{Semantics}: whether the model focuses on modeling special semantic gestures. \textit{Text-free}: whether the model works without text as input or intermediate modality. \textit{Hand}: whether hand modeling is considered. \textit{Motion}: motion generation approach.}
% \footnotesize
\setlength{\tabcolsep}{4pt}
\label{tab:related_work_eccv}
\begin{tabular}{lccccc}
\toprule
\textbf{Model} & \textbf{\makecell{Streaming?}} & \textbf{\makecell{Semantics?}} & \textbf{Text-free?} & \textbf{\makecell{Hand?}} & \textbf{\makecell{Motion?}} \\
\midrule
LivelySpeaker~\cite{Zhi_2023_ICCV} & \ding{55} & \ding{51} & \ding{55} & \ding{55} & Continuous \\
DiffSHEG~\cite{chen2024diffsheg} & \ding{51} & \ding{55} & \ding{51} & \ding{51} & Continuous \\
Semantic Gesticulator~\cite{zhang2024semantic}  & \ding{55} & \ding{51} & \ding{55} & \ding{51} & Discrete \\
SemTalk~\cite{zhang2025semtalk} & \ding{55} & \ding{51} & \ding{55} & \ding{51} & Discrete \\
\midrule
\textbf{Ours} & \ding{51} & \ding{51} & \ding{51} & \ding{51} & Continuous \\
\bottomrule
\end{tabular}
\vspace{2mm}
\vspace{-5mm}
\end{table*}

Gestures are integral nonverbal and non-manipulative body movements that enhance human communication~\cite{mcneill1992hand}. Research in co-speech gesture generation has transitioned from early rule-based~\cite{kipp2005gesture} and statistical machine learning methods~\cite{levine2010gesture} to modern deep learning frameworks~\cite{liu2025semges, chen2024diffsheg, yoon2020speech,zhang2025echomask,zhang2026mitigating,zhang2026personagesture,cheng2025holegest}, fueled by the availability of large-scale audio-gesture datasets~\cite{liu2022beat, ghorbani2023zeroeggs}.
As summarized in Table~\ref{tab:related_work_eccv}, recent efforts prioritize bridging gestures with linguistic content through diverse alignment strategies. LivelySpeaker~\cite{Zhi_2023_ICCV} utilizes CLIP~\cite{radford2021learning} to fuse rhythmic and semantic cues, yet it often fails to maintain semantic consistency. DiffSHEG~\cite{chen2024diffsheg} employs global semantic descriptors for alignment but lacks the precision required for fine-grained contexts. While Semantic Gesticulator~\cite{zhang2024semantic} achieves high-fidelity semantic correspondence via LLM-based retrieval, its dependence on post-hoc alignment hinders real-time streaming applications. Similarly, SemGes~\cite{liu2025semges} introduces coherence and relevance losses to ground semantics; however, its performance is constrained by sparse semantic annotations. SemTalk~\cite{zhang2025semtalk} attempts to decouple general and sparse motions, but its reliance on text-based features makes capturing precise temporal alignment with audio cues challenging.
To address these limitations, we propose an automatic, semi-synthetic data generation pipeline coupled with a two-stage training strategy. Our approach enables end-to-end synchronization between speech and expressive gestures by learning semantic cues directly from audio. This text-free design inherently supports low-latency streaming settings while ensuring robust cross-modal alignment.

\section{Method}
\label{sec:method}

% \subsection{Task Definition: Real-time Semantic-Aligned Gesture Generation}
% The primary objective is to enable humanoid robots to perform real-time, semantic-aligned, and kinematically safe co-speech gesture generation[cite: 1, 2]. Unlike offline social interaction animation, our task requires a streaming pipeline where the input is a continuous audio chunk $s_{human}$, and the output is a corresponding action chunk $m_{humanoid}$[cite: 2, 3]. 

% This setup presents a unique challenge: the robot must balance \textbf{low latency} for interactive deployment while overcoming the \textbf{modality eclipse} phenomenon—a state where the model's reliance on past kinematic inertia suppresses the influence of subtle audio and semantic cues[cite: 3, 10, 60, 62]. By addressing these, our framework can be seamlessly integrated with streaming LLMs for robust robot-human social interaction.
The primary objective of this work is to achieve \textbf{real-time, semantically aligned, and safety-aware} co-speech gesture generation that drives an interactive human–humanoid system on physical hardware. Our framework operates within a streaming-to-streaming paradigm: it processes incoming audio chunks and generates corresponding motions in real-time. This architecture ensures seamless integration with high-performance streaming Large Language Models and audio feeds, thereby facilitating natural human-robot social interaction.
For the physical embodiment, we employ the Unitree G1 humanoid robot integrated with BrainCo dexterous hands. Our motion generation focuses on upper-body dynamics, utilizing a kinematic state space of 41 degrees of freedom (DoFs): 17 DoFs for the upper torso and arms, and 24 DoFs for the dexterous hands. This configuration provides the requisite mechanical flexibility for generating expressive, human-like gestures.

% The humanoid motion DoFs are combined of 7 DoFs of global position and orientation, 29 DoFs of robot body joints and 24 DoFs of hand joints.

% The core objective of this work is to enable humanoid robots to interact with humans through both speech and gesture. Given human audio inputs $s_\text{human}$, the robot is required to generate corresponding audio outputs $s_\text{humanoid}$ and gesture motions $m_\text{humanoid}$, ensuring consistency between the two modalities. In practical deployment, robot output must be generated in a streaming manner to minimize interaction latency. To achieve this, we propose an integrated pipeline for model training (Sec \ref{sec:method}), collect additional audio-motion data to enhance the expressive capabilities of the robot (Sec \ref{sec:data}), and perform dedicated optimizations for real-world deployment (Sec \ref{sec:experiments}). We choose Unitree G1 robot equipped with Brainco Dexterous Hands as embodyment for our humanoid model (see at Figure \ref{fig:pipeline}), as it is popular among recent robotics research ~\cite{zhang2025track, ze2025twist, he2025asap}. The humanoid motion DoFs are combined of 7 DoFs of global position and orientation, 29 DoFs of robot body joints and 24 DoFs of hand joints.

\begin{figure*}[t]
\centering
  \includegraphics[width=0.95\textwidth]{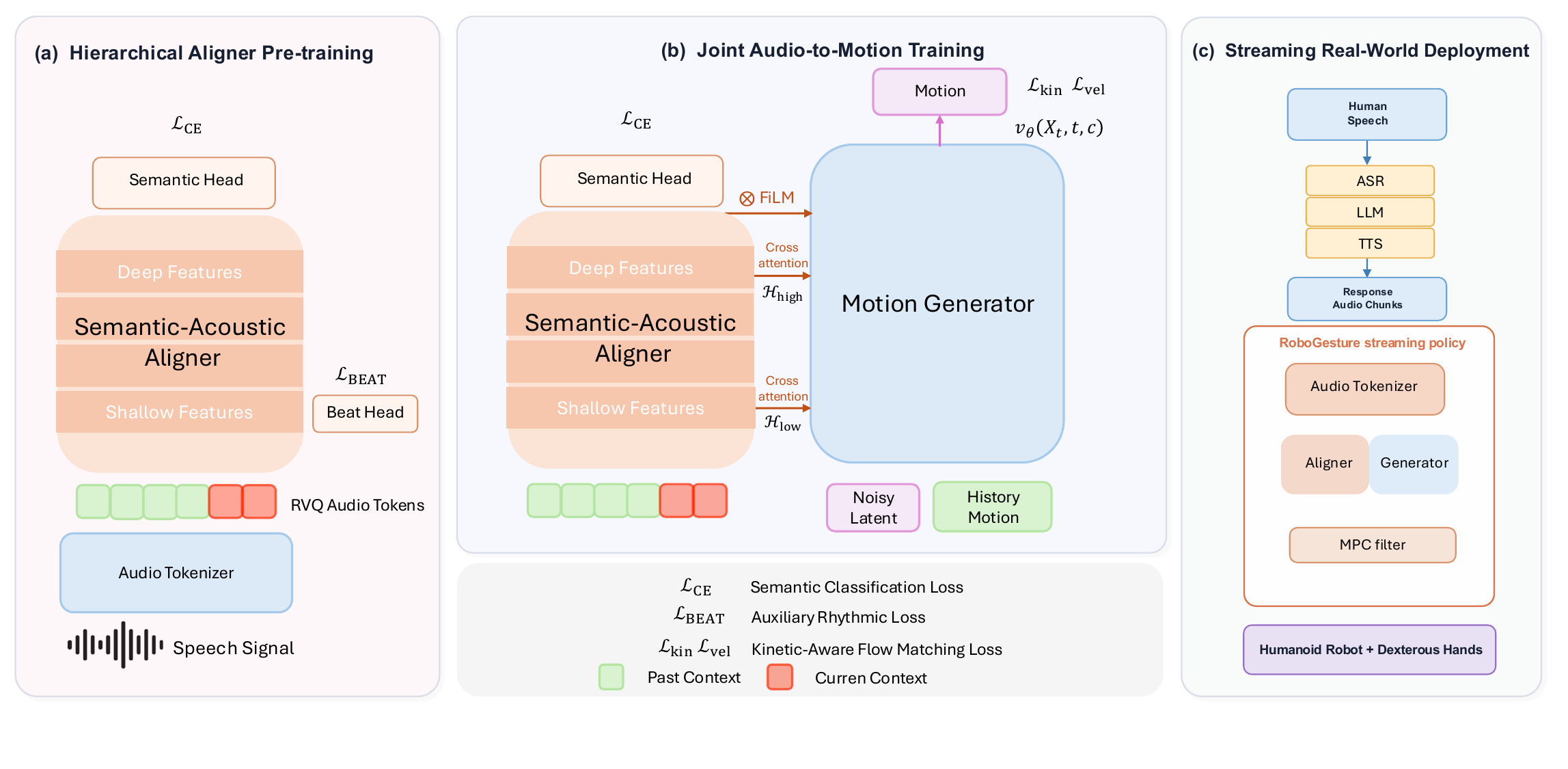}
  % \vspace{-1.5em}
  \caption{\textbf{Overview of RoboGesture.} \textbf{(a)} The Semantic-Acoustic Aligner is pre-trained to decouple multi-granular audio cues from streaming tokens. \textbf{(b)} The Motion Generator synthesizes motions via Conditional Flow Matching, jointly conditioned on hierarchical audio features and historical motion to maintain autoregressive consistency. \textbf{(c)} During inference, the pipeline seamlessly integrates with upstream Speech-LLMs and employs a safety filter to drive physical humanoid robots in real-time.}
\label{fig:training_pipeline}
% \vspace{-1em}
\end{figure*}

Our solution is a two-stage training framework structured into three core modules as shown in Figure~\ref{fig:training_pipeline}: (i) a hierarchical semantic-acoustic aligner, (ii) a streaming conditional motion generator, and (iii) an MPC-based kinematic safety filter.
Once trained, the framework enables seamless, end-to-end streaming from raw audio to robot trajectories. Section~\ref{sec:audio_rep}, ~\ref{sec:cfm_dit} and \ref{sec:mpc_refinement} describe the architectural details, while Section~\ref{sec:training} outlines the training objectives.

\subsection{Hierarchical Semantic-Acoustic Aligner}
\label{sec:audio_rep}
While recent approaches attempt to extract semantic cues from intermediate modalities such as text~\cite{zhang2025semtalk,zhang2024semantic,Zhi_2023_ICCV}, they often struggle to achieve precise temporal alignment between motion and subtle acoustic nuances. On one hand, converting audio to text inevitably leads to the loss of essential prosodic information, such as intonation and emphasis. For example, the word 'Really' conveys skepticism with a rising intonation but confirmation with a falling one—crucial distinctions embedded in raw audio that text-based methods fail to capture. On the other hand, co-speech gestures often exhibit pre-phonetic anticipation; for example, before articulating an emphatic verb like 'Stop!', the body typically initiates energy accumulation, such as leaning back or raising a hand. Such preparatory signals are hidden in the subtle acoustic precursors immediately preceding vocalization. To address these limitations, we propose to fully leverage multi-granular audio features through a semantic-acoustic aligner.

To get multi-level acoustic features, we utilize the Mimi codec~\cite{kyutai2024moshi} to tokenize the streaming audio. Unlike traditional codecs, Mimi explicitly decouples audio chunks into a hierarchical representation: the first quantizer provides high-level semantic tokens; while the subsequent residual quantizers capture fine-grained acoustic details such as prosody and intonation.

To harness these multi-granular cues, we introduce a transformer-based semantic-acoustic aligner optimized via multi-task auxiliary learning. Shallow layers are designed to capture rapid acoustic energy transients. We extract low-level features from these layers and route them to a beat head for auxiliary rhythmic supervision. This ensures the generated motion is precisely synchronized with acoustic onsets.  Conversely, deep layers distill macroscopic semantics via a 300-class discrete classification task powered by RoboGesture dataset in Section~\ref{sec:data}. By processing high-level features through a semantic head, the architecture acts as an information bottleneck, compressing acoustic nuances into actionable tokens that retain prosodic accuracy often lost in raw text.

Ultimately, this hierarchical disentanglement allows the aligner to provide distinct yet complementary control signals. The low-level rhythmic pulses and high-level semantic labels work in tandem to guide the downstream motion generator.
% ensuring both temporal precision and semantic depth.

\subsection{Streaming Conditional Motion Generator}
\label{sec:cfm_dit}

Existing \textit{Motion Tokenizer + Autoregressive (AR)} frameworks often struggle to capture high-fidelity kinematics, as discrete tokens lack the precision required for expressive hand gestures. Furthermore, AR models are prone to recursive error accumulation, compromising long-term semantic coherence. To overcome these limitations, we employ a Diffusion Transformer (DiT)\cite{peebles2023scalable} operating in a continuous state space, parameterized by Conditional Flow Matching (CFM)\cite{lipman2023flowmatchinggenerativemodeling}.

However, a critical challenge in this pipeline is the modality eclipse: the generator may rely excessively on historical kinematic inertia, producing motions that are plausible but audio-unaware. To ensure the gestures remain strictly driven by the streaming audio while maintaining physical continuity, we propose a dual condition injection mechanism: 
\textbf{(1) Framewise Micro-alignment:} We utilize cross-attention to fuse multi-scale acoustic cues with historical motion context. In this process, the shallow features that capture rhythmic transients and the deep features that provide local semantic guidance are extracted from the aligner. These multi-level sequences, alongside the \textit{Past Motion} chunk, are projected as Key-Value pairs. By explicitly injecting the historical kinematic state and granular acoustic signals through the attention mechanism, the model ensures seamless transitions and precise temporal synchronization across consecutive generation windows. 
\textbf{(2) Global Macro-modulation:} In contrast, the highly compressed semantic instructions derived from the aligner serve as macroscopic behavioral baselines. To prevent these global intents from being overshadowed by local kinematic noise, we employ Feature-wise Linear Modulation (FiLM) \cite{perez2018film}. FiLM applies affine transformations to the DiT's intermediate feature maps, robustly establishing the overarching emotional and semantic tone without interfering with intricate local alignments.

\subsection{MPC-based Collision-avoidance Filter}
\label{sec:mpc_refinement}

While training our framework directly within the robot's action space significantly mitigates the artifacts typical of manual retargeting, the stochastic nature of generative models cannot inherently guarantee absolute physical safety. 

To ensure collision-free and physically feasible deployment, we pass the synthesized motion chunks through an MPC-based optimization module in the inference time. This module acts as a safety filter by solving a constrained optimization problem in real-time. 
% It enforces collision constraints together with velocity and position tracking constraints, enabling fast optimization of existing trajectories. We detailed it in the supplementary material.
It enforces collision constraints together with velocity and position tracking constraints, which we cast as a convex quadratic program with velocity, trajectory-tracking, and smoothing costs, solved per frame with OSQP in a streaming manner. This filter reduces the self-collision frame ratio of the generated motions from 4.16\% to 0.13\%. We provide the full formulation and statistics in the supplementary material.

% \subsection{Training Recipes and Objectives}
% \label{sec:training_strategy}

\subsection{Training Recipes and Objectives}
\label{sec:training}

To develop a robust aligner and a highly responsive motion generator, we adopt a hierarchical two-stage training strategy as shown in Figure~\ref{fig:training_pipeline}(a)(b). Specifically, we define a temporal horizon of $T = 30$ frames for each one-second action chunk, operating in a robot action space of $D = 41$ dimensions. To incorporate sufficient context, we introduce a historical window of $T_{hist} = 2T$ frames.

\subsubsection{Stage 1: Representation Disentanglement Pre-training}
In the initial stage, the DiT generator remains frozen. We exclusively optimize the aligner using 1,000 hours of semantic-annotated chunk-level semi-synthetic data, as detailed in Section~\ref{sec:data}. To capture temporal context, the input consists of a $3T$-frame audio window, comprising the current chunk and a historical window. 
% This architecture forces the latent space to internalize multi-granular rhythmic and semantic knowledge via auxiliary supervision.
Intermediate features from the shallow layers are routed to a beat head. The rhythmic target $B_{target} \in \mathbb{R}^{T \times 2}$ is a concatenation of the normalized physical velocity magnitude from the ground-truth motion and the acoustic onset strength:
\begin{equation}
\begin{aligned}
    B_{target} = \big[&\text{Norm}(\text{Resample}(\| \dot{q} \|_2));\\[-0.2ex]
                      &\text{Norm}(\text{Onset}(wav))\big]
\end{aligned}
\end{equation}
where $\dot{q}$ represents the joint velocities. The deep layers are routed to a semantic head to predict discrete action categories $Y_{target}$. The Stage 1 objective is:
\begin{equation}
\begin{aligned}
    \mathcal{L}_{Stage1}
    ={}& \lambda_{beat}\,\text{MSE}(\hat{B}, B_{target}) \\
       &+ \text{CE}(\hat{Y}, Y_{target})
\end{aligned}
\end{equation}
where $\lambda_{beat}$ balances the regression and classification gradients.

\subsubsection{Stage 2: Joint Generative Training with CFG Masking}
In the second stage, we unfreeze the DiT to jointly optimize motion synthesis and semantic preservation. To balance expressive semantics with rhythmic alignment, we utilize a composite dataset blending the 1000-hour semantic-centric semi-synthetic data with an upsampled (3$\times$) beat-centric dataset (BEAT \cite{liu2022beat}, about 76 hours). To break the ``modality eclipse'', we introduce Anti-Inertia CFG Masking: the \textit{Past Motion} condition is randomly masked with a 15\% probability, compelling the network to perform ``cold-starts'' driven by aligner features.

To prevent fine-grained hand kinematics from being submerged, we define a spatial weight $W_{s} \in \mathbb{R}^D$ ($w_{hand}=4.0$ for hand joints) and a temporal weight $W_{t} \in \mathbb{R}^T$ ($w_{frame} \in \{5, 10\}$ for frames within semantic intervals). The core generative process is supervised by a spatio-temporally weighted Velocity Matching Loss:
\begin{equation}
\begin{aligned}
    \mathcal{L}_{vel}
    ={}& \mathbb{E}_{\tau,M_1,M_0}\Bigg[
        \sum_{t=1}^{T}\sum_{d=1}^{D} W_t^{(t)}W_s^{(d)} \\
       &\quad\cdot\Big(v_\theta(M_\tau,\tau,C)^{(t,d)}
        -(M_1-M_0)^{(t,d)}\Big)^2\Bigg]
\end{aligned}
\end{equation}
where $\tau \sim \mathcal{U}(0,1)$ is the flow matching time step, $M_0 \sim \mathcal{N}(0, I)$ is the standard Gaussian noise, $M_1$ is the target motion, and $M_\tau = (1-\tau)M_0 + \tau M_1$ is the intermediate interpolated state. To suppress high-frequency jittering, we explicitly penalize the first-order temporal difference. Using a single-step Euler approximation $\hat{M}_1 = M_\tau + (1-\tau)v_\theta(M_\tau, \tau, C)$, we introduce a Kinetic Consistency Loss ($\mathcal{L}_{kin}$):
\begin{equation}
\begin{aligned}
    \mathcal{L}_{kin}
    ={}& \mathbb{E}_{\tau,M_1,M_0}\Bigg[
        \sum_{t=1}^{T-1}\sum_{d=1}^{D}\bar{W}_t^{(t)}W_s^{(d)} \\
       &\quad\cdot\Big(\Delta\hat{M}_1^{(t,d)}
        -\Delta M_1^{(t,d)}\Big)^2\Bigg]
\end{aligned}
\end{equation}
where $\Delta M^{(t)} = M^{(t+1)} - M^{(t)}$ denotes the inter-frame displacement, and $\bar{W}_{t}^{(t)}$ is the averaged temporal weight of adjacent frames. Ultimately, the final joint objective for Stage 2 is:
\begin{equation}
\begin{aligned}
    \mathcal{L}_{Stage2}
    ={}& \mathcal{L}_{vel}+\lambda_{kin}\mathcal{L}_{kin} \\
       &+\lambda_{sem}\,\text{CE}(\hat{Y},Y_{target})
\end{aligned}
\end{equation}
where $\lambda_{kin}$ and $\lambda_{sem}$ are hyper-parameters that balance the trade-off.

\section{Automatic Semi-Synthetic Data Generation}
\label{sec:data}
% \blue{\lipsum[1]}

The size and quality of paired audio-motion data are critical for training multimodal models. Recent studies~\cite{zhang2024semantic, liu2025semges, chen2024diffsheg} typically rely on the BEAT dataset~\cite{liu2022beat} which contains about 76 h of motion-capture recordings. While BEAT is excellent for rhythm-synchronous learning, gestures that convey explicit semantics are extremely sparse, lying in the long-tail of natural human motion.
Semantic Gesticulator~\cite{zhang2024semantic} alleviates this issue through the SeG dataset, a curated collection of more than 200 semantic gesture classes. However, SeG still omits many everyday scenarios and contains interpenetration artifacts, which limits its suitability for real-world robot motion synthesis.

To address these limitations, we introduce an automatic semi-synthetic data generation pipeline that expands the scale and improves the quality of semantic gesture data paired with speech and text. Notably, the whole data generation pipeline is designed on humanoid motion representations, generating smooth and collision-free humanoid motions for native support of humanoid control.

\subsection{Retargeting and Collision-Avoidance}
\label{sec:4.1}
To obtain robot data that is feasible in the real world, two issues are central: transferring human motion to the robot while preserving accurate human-likeness, and ensuring that the transferred motions remain collision-free and executable for safe operation.

For accuracy, we use enhanced GMR~\cite{ze2025gmr} and Dex-Retargeting~\cite{qin2023anyteleop} to retarget body and hand motions to a G1 humanoid equipped with a customized BrainCo hand. We can handle human motions originating from different skeletal structures and enable accurate transfer of full-body and hand kinematics. For collision avoidance, we adopt the MPC-based filter in Section~\ref{sec:mpc_refinement} to clean the data.

\subsection{RoboGesture Dataset}

\begin{figure}[t]
\centering
  \includegraphics[width=1\columnwidth]{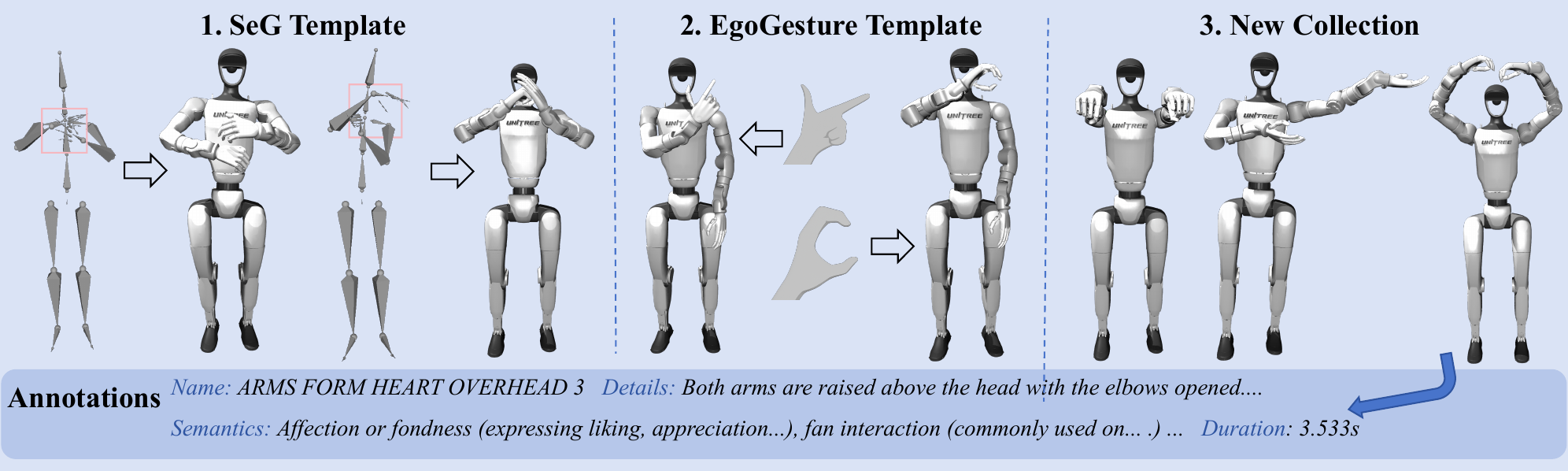}
  \caption{\textbf{Visualization of RoboGesture Dataset.} RoboGesture refines the SeG dataset by correcting motion penetration issues (e.g., avoiding arm–body collisions), augments the EgoGesture dataset with full-body motion, and additionally incorporates newly collected ones. Each sample is annotated with detailed semantic labels.}
\label{fig:robogesture_dataset}
% \vspace{-1.5em}
% \vspace{-0.6cm}
\end{figure}

As shown in Figure~\ref{fig:robogesture_dataset}, we build RoboGesture, a high-quality semantic gesture dataset for humanoid robots comprising over 300 gesture categories. The class list is derived from SeG~\cite{zhang2024semantic} and EgoGesture~\cite{zhang2018egogesture, cao2017egocentric} templates and expanded through user questionnaires.
Each gesture is recorded with a marker-based motion capture system to preserve fine-grained kinematics, and manually annotated with detailed motion descriptions and its possible meanings across different scenarios.

Each gesture is retargeted and optimized, and then replayed on the physical robot to verify reachability and control fidelity. The resulting RoboGesture dataset preserves human-like expressiveness while remaining fully executable on the robot, providing a reliable foundation for downstream gesture synthesis.

\subsection{Semi-Synthetic Data Generation Pipeline}

\label{sub:demo}

Based on RoboGesture, our retargeting pipeline and collision-avoidance tools, we generate audio–motion data enriched with semantic gestures.

Our synthesis pipeline comprises five stages: (1) Scenario Generation: We leverage LLMs~\cite{comanici2025gemini, openai2023gpt4} to produce diverse daily scenarios and narrative scripts enriched with emotional descriptions. (2) Gesture Tagging: LLMs select semantically appropriate gestures from RoboGesture and determine their optimal insertion points, adjusting the wording to ensure natural linguistic integration. (3) Multimodal Synthesis: We generate emotion-aware TTS audio with word-level timestamps alongside rhythmic beat gestures via a model trained on the BEAT dataset. (4) Temporal Blending: Following~\cite{zhang2024semantic, indefrey2004spatial}, semantic gestures are scheduled to initiate 0.4s before their corresponding spoken keywords—mimicking human anticipatory behavior—and are then fused with rhythmic motions using fifth-order interpolation. (5) Safety Optimization: Finally, we apply collision-avoidance refinement to the blended trajectories.
This automated, low-cost pipeline enables the generation of millions of high-quality, semi-synthetic audio-motion pairs across diverse conversational contexts by simply varying the input prompts or text libraries.

\section{Experiments}
\label{sec:experiments}

To assess gesture quality, we first evaluate on the standard co-speech gesture generation task. This task, typically conducted in simulation, requires the model to produce natural and synchronized gesture motions conditioned on  speech and serves as a widely-used benchmark with established metrics.

To further evaluate the model's interactive capabilities and potential for real-world deployment, we conduct experiments on a social interaction task in a physical setting. In this task, given an audio input, the model is required to generate expressive and physically safe multimodal responses that encompass both audio and motion in a streaming manner.

\subsection{Evaluating on Co-speech Gesture Generation}

\noindent\textbf{Datasets.}
(1) \textbf{BEAT.} The BEAT dataset~\cite{liu2022beat} comprises 76 hours of multimodal recordings from 30 participants, encompassing speech audio, transcriptions, and high-quality motion capture data. It serves as a standard benchmark for training and evaluating co-speech gesture generation models and includes semantic labels for quantitative evaluation.
(2) \textbf{SemanticBEAT.} The BEAT dataset exhibits limited domain coverage, and its semantic gesture annotations are often vague. These factors render it insufficient for the rigorous evaluation of semantic-aware co-speech gesture generation. To address these limitations, we introduce \emph{SemanticBEAT}, a new test set comprising 1,000 speech videos compiled from online sources and user surveys. For this dataset, we performed manual temporal annotation of semantic gestures, precisely marking the intervals during which speakers should perform meaningful gestures.

\noindent\textbf{Baselines. } We compare our method with recent state-of-the-art baselines that provide pretrained checkpoints (trained on BEAT). 
(1) \textbf{LivelySpeaker} (ICCV 2023)~\cite{Zhi_2023_ICCV}: generates semantically and rhythmically aware gestures using an MLP-based diffusion model. Since it produces fixed-length motion clips (34 frames), we follow the authors’ strategy to smoothly concatenate consecutive clips.
(2) \textbf{Diffsheg} (CVPR 2024)~\cite{chen2024diffsheg}: aligns gestures with audio/text using global semantic cues.
(3) \textbf{SemTalk} (ICCV 2025)~\cite{zhang2025semtalk}: separately learn general motions and sparse motions with adaptive fusion.
(4) \textbf{Semantic Gesticulator(SG)} (SIGGRAPH 2024)~\cite{zhang2024semantic}: first produces rhythmic gestures and then refines semantic gestures via an LLM-based retrieval module.
All baseline outputs are originally human motions; thus, we apply meticulous retargeting tuned to the robot to ensure a fair comparison of motion quality. We further verify in the supplementary material that re-training representative baselines directly in the robot joint space yields nearly identical results, confirming that our gains stem from the model rather than from retargeting artifacts.

\subsubsection{Quantitative Evaluation} 
\noindent{\textbf{Metrics.}} Following~\cite{zhang2024semantic,Zhi_2023_ICCV,chen2024diffsheg,liu2025semges}, we evaluate our method using five key metrics: 
(1) \textbf{Fréchet Gesture Distance (FGD)}~\cite{yoon2020speech} evaluates the fidelity of generated gestures to the real motion distribution by embedding sequences into a latent space via a pre-trained autoencoder. 
(2) \textbf{Beat Consistency (BC)}~\cite{li2021learn} quantifies speech-motion synchronization by measuring the alignment between audio onsets and motion beats (velocity minima in upper-body joints). 
(3) \textbf{Diversity (DIV)}~\cite{li2021audio2gestures} assesses the variability of generated motions, computed as the average $L_1$ distance between pairs of $N$ generated clips. 
(4) \textbf{Collision Rate (Col.)} measures the physical plausibility of the motion; we employ GMR~\cite{ze2025gmr} to detect self-collisions between the robot's mesh components. 
(5) \textbf{Mean Squared Error (MSE)}~\cite{ijcai2023p650} calculates the average Euclidean distance between the generated joint positions and the ground-truth sequences to measure structural reconstruction accuracy. Metrics (1)(5) were omitted for SemanticBeat as it is an out-of-domain test set lacking ground truth. 
% All metrics are calculated on gestures mapped to the robot's morphology, as detailed in the supplementary material.

\begin{table*}[t]
\centering
\scriptsize
% 确保导言区有 \usepackage{booktabs} 和 \usepackage{graphicx}
\caption{Quantitative comparison of our model with other methods on the BEAT and SemanticBEAT datasets. Since the SemanticBEAT dataset does not provide ground-truth motion or semantic annotations, we omit FGD and MSE for this dataset. Best results are shown in \textbf{bold}, and second-best results are \underline{underlined}.}
% \vspace{-0.5em}
\label{tab:main_table}
\begin{tabular}{l c c c c c | c c c}
\toprule
& \multicolumn{5}{c|}{\textbf{BEAT}} & \multicolumn{3}{c}{\textbf{SemanticBEAT}} \\
\cmidrule(r){2-6} \cmidrule(l){7-9}
\textbf{Method} & \textbf{FGD $\downarrow$} & \textbf{BC $\uparrow$} & \textbf{MSE $\downarrow$} & \textbf{DIV $\uparrow$} & \textbf{Col. $\downarrow$} & \textbf{BC $\uparrow$}  & \textbf{DIV $\uparrow$} & \textbf{Col. $\downarrow$} \\
\midrule
LivelySpeaker~\cite{Zhi_2023_ICCV} & 3.0350 & 0.1811 & 0.1861 & 0.1485 & 21.41 & 0.2852 & 0.1384 & 13.36 \\
DiffSHEG~\cite{chen2024diffsheg} & \underline{2.2316} & \underline{0.1851} & \underline{0.1753} & 0.1218 & \textbf{0.85} & 0.2859 & 0.1209 & \underline{1.20} \\
SemTalk~\cite{zhang2025semtalk} & 7.9328 & 0.1828 & 0.3931 & 0.1781 & 52.82 & \underline{0.2913} & 0.1440 & 42.65 \\
Semantic Gesticulator~\cite{zhang2024semantic} & 3.0147 & 0.1771 & 0.2375 & \textbf{0.2818} & 13.11 & 0.2906 & \textbf{0.2831} & 13.84 \\
\midrule
Ours & \textbf{0.8452} & \textbf{0.1866} & \textbf{0.1347} & \underline{0.2075} & \underline{0.88} & \textbf{0.2950} & \underline{0.2041} & \textbf{0.13} \\
\bottomrule
\end{tabular}
% \vspace{-1em}
\end{table*}

\begin{figure*}[t]
  \centering
  \includegraphics[width=0.9\textwidth]{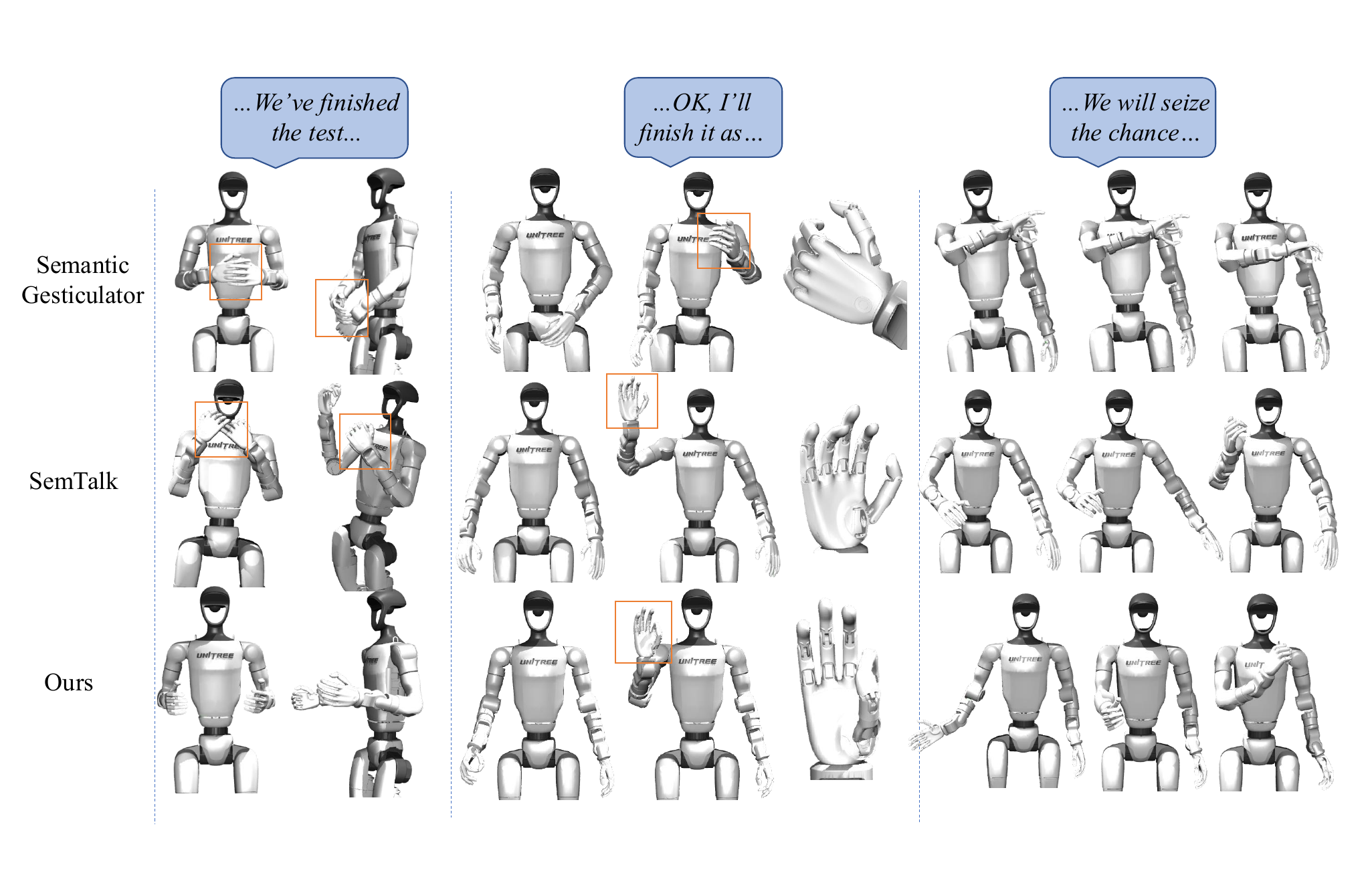}
  \caption{\textbf{Qualitative comparison of generated co-speech gestures.} We present three representative scenarios to highlight the advantages of our method. \textbf{Left:} Baseline methods (e.g., SG) occasionally suffer from severe self-penetration (highlighted by the orange box), whereas our model ensures physical safety and strictly avoids self-collisions. \textbf{Middle:} Given the explicit semantic cue ``OK'', our model successfully synthesizes the precise, fine-grained ``OK'' hand gesture, while baselines fail to capture this specific semantic alignment. \textbf{Right:} For emphatic speech contexts (``seize the chance''), our model generates highly expressive body language, such as confidently patting the chest, demonstrating superior contextual richness and overall expressiveness.}
  \label{fig:exp_cospeech}
  % \vspace{-2em}
\end{figure*}
\begin{table*}[t]  
    \centering
    \scriptsize
    \caption{Human Evaluation on BEAT and SemanticBEAT datasets. We evaluate models across four key dimensions: Rhythmic Alignment (RA), Semantic Accuracy (SA), Physical \& Hand Consistency (PHC), and Overall Preference (OP). All values are reported as mean $\pm$ standard error. Higher values indicate better performance. }
    % \vspace{-1em}
    \label{tab:comparison_results}
    \begin{tabular}{l cccc cccc}
        \toprule
        & \multicolumn{4}{c}{\textbf{BEAT}} & \multicolumn{4}{c}{\textbf{SemanticBEAT}} \\
        \cmidrule(lr){2-5} \cmidrule(lr){6-9}
        \textbf{Model} & RA$\uparrow$ & SA$\uparrow$ & PHC$\uparrow$ & OP$\uparrow$ & RA$\uparrow$ & SA$\uparrow$ & PHC$\uparrow$ & OP$\uparrow$ \\
        \midrule
        LivelySpeaker~\cite{Zhi_2023_ICCV} & -0.2223 & -0.3994 & -0.4304 & -0.3064 & -0.2126 & -0.3015 & -0.3466 & -0.3497 \\
        Diffsheg~\cite{chen2024diffsheg} & -0.0058 & -0.2691 & -0.1440 & -0.0992 & -0.0090 & -0.0817 & -0.0458 & -0.0626 \\
        SemTalk~\cite{zhang2025semtalk} & 0.0128 & 0.0151 & 0.1679 & -0.0132 & 0.0650 & -0.0012 & 0.1207 & 0.0970 \\
        SemanticGesticulator~\cite{zhang2024semantic} & 0.0819 & 0.2264 & -0.0273 & 0.1137 & 0.0296 & 0.1605 & -0.0347 & 0.0640 \\
        \midrule
        \textbf{Ours} & \textbf{0.1334} & \textbf{0.4270} & \textbf{0.4338} & \textbf{0.3050} & \textbf{0.1269} & \textbf{0.2239} & \textbf{0.3066} & \textbf{0.2514} \\
        \bottomrule
    \end{tabular}
    % \vspace{-1em}
\end{table*}

\noindent\textbf{Results.} Table~\ref{tab:main_table} demonstrates that our method outperforms baselines across most metrics. We achieve state-of-the-art FGD/BC/MSE on the BEAT benchmark, indicating superior alignment with the ground-truth distribution and accurate structural reconstruction. Furthermore, our model ensures strict physical plausibility with a near-zero Collision Rate (0.88\%/0.13\%). While SG yields higher DIV scores, numerical diversity can be artificially inflated by unstable or jittery motions~\cite{chen2024diffsheg}. Overall, our approach strikes an optimal balance between kinematic realism, speech-motion synchronization, and safe gesture diversity. Figure~\ref{fig:exp_cospeech} provides the corresponding qualitative comparison.

\subsubsection{Human Evaluation.}
To achieve a more comprehensive assessment of semantic-aligned, safe, and human-like motion qualities, we conducted a subjective user study based on pairwise comparisons. Following~\cite{parizet2005comparison,alexanderson2023listen,zhang2024semantic}, We randomly sampled 20 generated gesture clips for each baseline, yielding a total of 200 comparison pairs. Participants were asked to evaluate the paired motions across four key dimensions: Rhythmic Alignment, Semantic Accuracy, Physical \& Hand Consistency, and Overall Preference . The pairwise responses were then aggregated into merit scores to reflect relative performance. Further details regarding the user study setup and score calculation are provided in the supplementary material.

\noindent\textbf{Results.} The subjective evaluation results are summarized in Table~\ref{tab:comparison_results}. Our method consistently outperforms all baselines across all dimensions. Specifically, our model achieves the highest scores in Physical \& Hand Consistency, verifying that our approach successfully suppresses high-frequency jittering and generates safer, physically plausible motions. Furthermore, our substantial lead in Rhythmic Alignment and Semantic Accuracy demonstrates that our framework produces gestures that are not only temporally synchronized with audio beats but also highly contextually appropriate. Consequently, our method secures the highest Overall Preference, confirming that the generated trajectories are perceived as significantly more natural and human-like compared to existing baselines.

% 使用200人进行打分
% 解释没有semi-data失去了手部丰富指向明确的语义动作 只有节拍动作 1/4即200h半合成数据表现力削减 在结构上在crossattention减少pastmotion输入会使动作跳变 减少了FiLM会使语义准确性下降/感染力不足 去掉分类任务只模态遮蔽模型不从audiotoken学习语义信息会使准确率大幅度下降 CFG可以实现避免motion只从past状态推理得出，促进跨模态特征的均衡接纳 AR的模型语义准确性和动作还原度/流畅度都存在问题 失去Kinetic-Aware (KA) Loss减少了对手部动作，语义区间和动作平滑度的监督造成表现力下降 最后 没有filter导致平滑度略有损失
\subsection{Ablation Study}
\label{sec:ablation}

\noindent\textbf{Experimental Setup.} To thoroughly evaluate the contribution of each proposed component in our framework, we conduct a comprehensive ablation study. We recruited 200 participants to evaluate the generated motions across four key dimensions: Semantic Action Score, Hand Detail Score, Human-likeness \& Naturalness, and Beat Matching Score. The quantitative results of these configurations are reported in Table~\ref{tab:ablation_study}. We group our analysis into three main perspectives: data scale, architecture and strategy, and loss refinement. 

\begin{table}[!t]
    \centering
    % 在 Caption 中详细解释缩写
    \caption{Ablation study on the key components of our method. SA: Semantic Action Score, HD: Hand Detail Score, HN: Human-likeness \& Naturalness, BM: Beat Matching Score. Bold indicates the best performance among all configurations.}
    \label{tab:ablation_study}
    \scriptsize
    \setlength{\tabcolsep}{2.2pt}
    % The ablation table is deliberately single-column so that it can stay
    % beside the ablation discussion instead of drifting into References.
    \begin{tabular*}{\columnwidth}{@{\extracolsep{\fill}}l cccc}
        \toprule
        \textbf{Configuration} & SA $\uparrow$ & HD $\uparrow$ & HN $\uparrow$ & BM $\uparrow$ \\
        \midrule
        \rowcolor[gray]{.95} \multicolumn{5}{l}{\textit{Data Scale}} \\
        Ours w/o Semi-data                & 4.281 & 2.321 & 4.945 & 4.628 \\
        Ours (1/4 Semi-data)              & 6.316 & 5.980 & 6.882 & 6.892 \\
        \midrule
        \rowcolor[gray]{.95} \multicolumn{5}{l}{\textit{Architecture \& Strategy}} \\
        Ours w/o Context Motion           & 4.237 & 4.263 & 2.192 & 1.928 \\
        Ours w/o FiLM Injection           & 5.181 & 4.389 & 4.506 & 4.589 \\
        Ours w/o Semantic Classification  & 5.007 & 4.747 & 4.750 & 5.268 \\
        Ours w/o CFG                      & 4.628 & 4.885 & 6.453 & 6.212 \\
        AR Strategy                       & 4.843 & 5.031 & 5.358 & 6.850 \\
        \midrule
        \rowcolor[gray]{.95} \multicolumn{5}{l}{\textit{Loss \& Refinement}} \\
        Ours w/o Kinetic-Aware (KA) Loss  & 5.573 & 3.947 & 4.763 & 5.587 \\
        Ours w/o Filter                   & 6.541 & 6.913 & 6.891 & 7.387 \\
        \midrule
        \textbf{Ours (Full Model)} & \textbf{7.175} & \textbf{7.203} & \textbf{7.394} & \textbf{7.529} \\
        \bottomrule
    \end{tabular*}
    % \vspace{-1.5em}
\end{table}

\noindent\textbf{Data Scale:} The semi-synthetic dataset plays a vital role in our system. Removing it entirely (\textit{Ours w/o Semi-data}) causes a catastrophic drop in SA and HD, as the model loses the ability to generate rich, directionally explicit semantic hand gestures, degrading to producing merely rhythmic beat motions. Even when trained on a subset (\textit{Ours 1/4 Semi-data}, approx. 250 hours), the overall expressiveness and semantic richness remain noticeably sub-optimal.

\noindent\textbf{Architecture \& Strategy:} Discarding the past motion context in the cross-attention module (\textit{Ours w/o Context Motion}) leads to severe motion discontinuities and jumps, reflected in the lowest HN and BM scores. Removing the FiLM injection (\textit{Ours w/o FiLM Injection}) impairs the model's expressiveness and semantic accuracy. Furthermore, omitting the semantic classification task (\textit{Ours w/o Semantic Classification}) prevents the masked model from explicitly learning semantic cues from audio tokens, resulting in a significant accuracy drop. Classifier-Free Guidance (\textit{Ours w/o CFG}) is also crucial; without it, the model tends to infer motions solely from past states rather than maintaining a balanced integration of cross-modal features.

\noindent\textbf{Loss \& Refinement:} The Kinetic-Aware Loss provides essential supervision. Removing it (\textit{Ours w/o KA Loss}) weakens the constraints on hand movements, semantic intervals, and kinematic smoothness, leading to a profound degradation in expressiveness. Finally, disabling the post-processing filter (\textit{Ours w/o Filter}) results in a slight but noticeable penalty in HN, confirming its necessity for ensuring ultimate motion smoothness.

\subsection{Evaluating Real-world Social Interaction}

Beyond gesture generation alone, we instantiate a complete interactive human–humanoid system in which the robot listens, responds, and gestures in real time. As detailed in the supplementary material, the deployed system comprises three modules: a language interaction module (ASR, a LoRA-tuned LLM, and TTS), our streaming speech-to-motion module (the audio aligner and motion generator), and a robot execution module. By leveraging our streaming aligner and motion generator, this pipeline achieves low-latency, end-to-end social interaction between humans and humanoid robots, as shown in 
% By leveraging our streaming audio aligner and motion generator, our framework enables low-latency, end-to-end social interaction between humans and humanoid robots. The pipeline interfaces with streaming audio foundation models or LLMs paired with streaming Text-to-Speech systems as shown in 
Figure~\ref{fig:training_pipeline}(c), which is then processed by our model to generate synchronized motion chunks. 

% The total system latency is primarily determined by the duration of the first generated audio chunk (typically $< 1.5$\,s) and the first-chunk inference time ($\approx 0.25$\,s).
The total system latency is primarily determined by the duration of the first generated audio chunk (typically $< 1.5$\,s) and the first-chunk inference time ($\approx 0.25$\,s). Crucially, the motion stack itself runs well above real time: the streaming motion generator sustains $\approx 120$\,FPS ($\approx 0.25$\,s per 1-second chunk) and the MPC safety filter adds only 5.6 ms per frame, both comfortably exceeding the 30 Hz robot control rate. The dominant latency therefore stems from the upstream speech pipeline (ASR, LLM, and TTS) rather than from our motion model; a full per-module latency breakdown is provided in the supplementary material.
Our experimental platform consists of a Unitree G1 robot equipped with dual BrainCo dexterous hands, utilizing a Proportional-Derivative (PD) control scheme to translate high-level trajectories into low-level motor commands. As illustrated in Figure \ref{fig:eccv_teaser}, the experimental results validate the framework's capability to generate semantically coherent and safety-constrained gestures. 

% To guarantee safety at real-world applications, we add an additional MPC framework before sending humanoid control signals to a real humanoid robot. Specifically, we formulate this problem as convex quadratic programming(QP), where the optimization goal includes velocity regularization and tracking error minimization, and the constraint is keeping the distance from obstacles above a certain threshould. To linearize the collision constraint, the Mujoco platform is used to calculate the contact jacobian matrix. 

% Our experiment setup includes a Unitree G1 robot with two Brainco Dexterous hands. We use PD control to send the low-level command to the robot controller. The experiment result is shown in Figure \ref{fig:teaser}, which demonstrates our ability in generating semantically appropriate and safe gestures. More details are provided in supplementary.

\section{Conclusion}

In this work, we present a robot-centric framework for enabling humanoid robots to engage in natural, real-time multimodal interaction through synchronized speech and expressive gestures. Our approach co-designs data curation, model architecture, and safety mechanisms with three main contributions: (1) a holistic framework integrating data, modeling, and control; (2) a data synthesis pipeline that produces a large-scale, semantically rich training corpus by mapping expressive human gestures to robot-specific kinematics; and (3) A continuous audio-driven motion policy to ensure tight temporal synchronization and prevent the model from collapsing into repetitive historical motion patterns.
Through experiments on a humanoid robot, we demonstrate that our system generates more semantically appropriate and safety-aware responses compared to baselines.

\section{Acknowledgments}
We thank Galbot for its generous support of this project. We are also
grateful to Benhao Qin and colleagues for their assistance with the
hardware deployment.

% This work provides a foundation for future research in conversational humanoid robotics with applications in assistive care, education, and collaboration.

% Our work has limitations that point to future directions. Precise temporal alignment between gestures and speech remains challenging, and we lack comprehensive quantitative metrics for evaluating fine-grained audio-motion alignment in humanoid contexts. The instruction-following module can be further fine-tuned to better suit real-world deployment scenarios through domain-specific tuning and more extensive interaction data collection.

% Do not allow unresolved manuscript floats to enter the reference list.
\FloatBarrier
\bibliographystyle{assets/plainnat}
\bibliography{main}

@String(CVPR= {IEEE Conf. Comput. Vis. Pattern Recog.})

@String(ICCV= {Int. Conf. Comput. Vis.})

@String(TOG= {ACM Trans. Graph.})

@String(AAAI = {AAAI})

@article{matheus2025long,
  title={Long-Term Interactions with Social Robots: Trends, Insights, and Recommendations},
  author={Matheus, Kayla and Ramnauth, Rebecca and Scassellati, Brian and Salomons, Nicole},
  journal={ACM Transactions on Human-Robot Interaction},
  volume={14},
  number={3},
  pages={1--42},
  year={2025},
  publisher={ACM New York, NY}
}

@article{dafarra2024icub3,
  title={icub3 avatar system: Enabling remote fully immersive embodiment of humanoid robots},
  author={Dafarra, Stefano and Pattacini, Ugo and Romualdi, Giulio and Rapetti, Lorenzo and Grieco, Riccardo and Darvish, Kourosh and Milani, Gianluca and Valli, Enrico and Sorrentino, Ines and Viceconte, Paolo Maria and others},
  journal={Science Robotics},
  volume={9},
  number={86},
  pages={eadh3834},
  year={2024},
  publisher={American Association for the Advancement of Science}
}

@inproceedings{yoon2019robots,
  title={Robots learn social skills: End-to-end learning of co-speech gesture generation for humanoid robots},
  author={Yoon, Youngwoo and Ko, Woo-Ri and Jang, Minsu and Lee, Jaeyeon and Kim, Jaehong and Lee, Geehyuk},
  booktitle={2019 International Conference on Robotics and Automation (ICRA)},
  pages={4303--4309},
  year={2019},
  organization={IEEE}
}

@inproceedings{mascaro2024robot,
  title={Robot interaction behavior generation based on social motion forecasting for human-robot interaction},
  author={Mascaro, Esteve Valls and Yan, Yashuai and Lee, Dongheui},
  booktitle={2024 IEEE International Conference on Robotics and Automation (ICRA)},
  pages={17264--17271},
  year={2024},
  organization={IEEE}
}

@article{valls2025robot,
  title={Robot Behavior Generation for Social Human-Robot Interaction},
  author={Valls Mascaro, Esteve and Lee, Dongheui},
  journal={International Journal of Social Robotics},
  pages={1--20},
  year={2025},
  publisher={Springer}
}

@inproceedings{zhang2025echomask,
  title={Echomask: Speech-queried attention-based mask modeling for holistic co-speech motion generation},
  author={Zhang, Xiangyue and Li, Jianfang and Zhang, Jiaxu and Ren, Jianqiang and Bo, Liefeng and Tu, Zhigang},
  booktitle={Proceedings of the 33rd ACM International Conference on Multimedia},
  pages={10827--10836},
  year={2025}
}

@article{galatolo2025simultaneous,
  title={Simultaneous text and gesture generation for social robots with small language models},
  author={Galatolo, Alessio and Winkle, Katie},
  journal={Frontiers in Robotics and AI},
  volume={12},
  pages={1581024},
  year={2025},
  publisher={Frontiers Media SA}
}

@inproceedings{zhang2026mitigating,
  title={Mitigating error accumulation in co-speech motion generation via global rotation diffusion and multi-level constraints},
  author={Zhang, Xiangyue and Li, Jianfang and Ren, Jianqiang and Zhang, Jiaxu},
  booktitle={Proceedings of the AAAI Conference on Artificial Intelligence},
  volume={40},
  pages={12834--12842},
  year={2026}
}

@inproceedings{wang2025mobileh2r,
  title={MobileH2R: Learning Generalizable Human to Mobile Robot Handover Exclusively from Scalable and Diverse Synthetic Data},
  author={Wang, Zifan and Chen, Ziqing and Chen, Junyu and Wang, Jilong and Yang, Yuxin and Liu, Yunze and Liu, Xueyi and Wang, He and Yi, Li},
  booktitle={Proceedings of the Computer Vision and Pattern Recognition Conference},
  pages={17315--17325},
  year={2025}
}

@inproceedings{wang2024genh2r,
  title={GenH2R: Learning generalizable human-to-robot handover via scalable simulation demonstration and imitation},
  author={Wang, Zifan and Chen, Junyu and Chen, Ziqing and Xie, Pengwei and Chen, Rui and Yi, Li},
  booktitle={Proceedings of the IEEE/CVF Conference on Computer Vision and Pattern Recognition},
  pages={16362--16372},
  year={2024}
}

@article{zhang2026personagesture,
  title={PersonaGesture: Single-Reference Co-Speech Gesture Personalization for Unseen Speakers},
  author={Zhang, Xiangyue and Cai, Yiyi and Li, Kunhang and Yang, Kaixing and Zhou, You and Li, Zhengqing and Chu, Xuangeng and Zhang, Jiaxu and Liu, Haiyang},
  journal={arXiv preprint arXiv:2605.06064},
  year={2026}
}

@article{cao2024ai,
  title={Ai robots and humanoid ai: Review, perspectives and directions},
  author={Cao, Longbing},
  journal={arXiv preprint arXiv:2405.15775},
  year={2024}
}

@inproceedings{jiang2025solami,
  title={Solami: Social vision-language-action modeling for immersive interaction with 3d autonomous characters},
  author={Jiang, Jianping and Xiao, Weiye and Lin, Zhengyu and Zhang, Huaizhong and Ren, Tianxiang and Gao, Yang and Lin, Zhiqian and Cai, Zhongang and Yang, Lei and Liu, Ziwei},
  booktitle={Proceedings of the Computer Vision and Pattern Recognition Conference},
  pages={26887--26898},
  year={2025}
}

@article{
  Ao2023GestureDiffuCLIP,
  author = {Ao, Tenglong and Zhang, Zeyi and Liu, Libin},
  title = {GestureDiffuCLIP: Gesture Diffusion Model with CLIP Latents},
  journal = {ACM Trans. Graph.},
  issue_date = {August 2023},
  numpages = {18},
  doi = {10.1145/3592097},
  publisher = {ACM},
  address = {New York, NY, USA},
  year = {2023}
}

@inproceedings{ijcai2023p650,
  title     = {DiffuseStyleGesture: Stylized Audio-Driven Co-Speech Gesture Generation with Diffusion Models},
  author    = {Yang, Sicheng and Wu, Zhiyong and Li, Minglei and Zhang, Zhensong and Hao, Lei and Bao, Weihong and Cheng, Ming and Xiao, Long},
  booktitle = {Proceedings of the Thirty-Second International Joint Conference on
               Artificial Intelligence, {IJCAI-23}},
  publisher = {International Joint Conferences on Artificial Intelligence Organization},
  pages     = {5860--5868},
  year      = {2023},
  month     = {8},
  doi       = {10.24963/ijcai.2023/650},
  url       = {https://doi.org/10.24963/ijcai.2023/650},
}

@InProceedings{Cai_2024_CVPR,
    author    = {Cai, Zhongang and Jiang, Jianping and Qing, Zhongfei and Guo, Xinying and Zhang, Mingyuan and Lin, Zhengyu and Mei, Haiyi and Wei, Chen and Wang, Ruisi and Yin, Wanqi and Pan, Liang and Fan, Xiangyu and Du, Han and Gao, Peng and Yang, Zhitao and Gao, Yang and Li, Jiaqi and Ren, Tianxiang and Wei, Yukun and Wang, Xiaogang and Loy, Chen Change and Yang, Lei and Liu, Ziwei},
    title     = {Digital Life Project: Autonomous 3D Characters with Social Intelligence},
    booktitle = {Proceedings of the IEEE/CVF Conference on Computer Vision and Pattern Recognition (CVPR)},
    month     = {June},
    year      = {2024},
    pages     = {582-592}
}

@article{yang2025omniretarget,
  title={OmniRetarget: Interaction-Preserving Data Generation for Humanoid Whole-Body Loco-Manipulation and Scene Interaction},
  author={Yang, Lujie and Huang, Xiaoyu and Wu, Zhen and Kanazawa, Angjoo and Abbeel, Pieter and Sferrazza, Carmelo and Liu, C Karen and Duan, Rocky and Shi, Guanya},
  journal={arXiv preprint arXiv:2509.26633},
  year={2025}
}

@article{joao2025gmr,
  title={Retargeting Matters: General Motion Retargeting for Humanoid Motion Tracking},
  author= {Joao Pedro Araujo and Yanjie Ze and Pei Xu and Jiajun Wu and C. Karen Liu},
  year= {2025},
  journal= {arXiv preprint arXiv:2510.02252}
}

@inproceedings{bettosi2024systematic,
  title={A Systematic Approach to Modeling Structured Behavior in Social Robots},
  author={Bettosi, Carl and Baillie, Lynne and Ross, Martin K and Broz, Frank},
  booktitle={Proceedings of the 2024 International Symposium on Technological Advances in Human-Robot Interaction},
  pages={29--37},
  year={2024}
}

@inproceedings{scherf2024learning,
  title={Learning action conditions for automatic behavior tree generation from human demonstrations},
  author={Scherf, Lisa and Fr{\"o}hlich, Kevin and Koert, Dorothea},
  booktitle={Companion of the 2024 ACM/IEEE International Conference on Human-Robot Interaction},
  pages={950--954},
  year={2024}
}

@inproceedings{tagliamonte2024generalizable,
  title={A generalizable architecture for explaining robot failures using behavior trees and large language models},
  author={Tagliamonte, Christian and Maccaline, Daniel and LeMasurier, Gregory and Yanco, Holly A},
  booktitle={Companion of the 2024 ACM/IEEE International Conference on Human-Robot Interaction},
  pages={1038--1042},
  year={2024}
}

@inproceedings{chen2024diffsheg,
  title={Diffsheg: A diffusion-based approach for real-time speech-driven holistic 3d expression and gesture generation},
  author={Chen, Junming and Liu, Yunfei and Wang, Jianan and Zeng, Ailing and Li, Yu and Chen, Qifeng},
  booktitle={Proceedings of the IEEE/CVF Conference on Computer Vision and Pattern Recognition},
  pages={7352--7361},
  year={2024}
}

@book{kipp2005gesture,
  title={Gesture generation by imitation: From human behavior to computer character animation},
  author={Kipp, Michael},
  year={2005},
  publisher={Universal-Publishers}
}

@book{mcneill1992hand,
  title={Hand and mind: What gestures reveal about thought},
  author={McNeill, David},
  year={1992},
  publisher={University of Chicago press}
}

@incollection{levine2010gesture,
  title={Gesture controllers},
  author={Levine, Sergey and Kr{\"a}henb{\"u}hl, Philipp and Thrun, Sebastian and Koltun, Vladlen},
  booktitle={Acm siggraph 2010 papers},
  pages={1--11},
  publisher={ACM},
  year={2010}
}

@article{yoon2020speech,
  title={Speech gesture generation from the trimodal context of text, audio, and speaker identity},
  author={Yoon, Youngwoo and Cha, Bok and Lee, Joo-Haeng and Jang, Minsu and Lee, Jaeyeon and Kim, Jaehong and Lee, Geehyuk},
  journal={ACM Transactions on Graphics (TOG)},
  volume={39},
  number={6},
  pages={1--16},
  year={2020},
  publisher={ACM New York, NY, USA}
}

@article{parizet2005comparison,
  title={Comparison of some listening test methods: a case study},
  author={Parizet, Etienne and Hamzaoui, Nacer and Sabatie, Guillaume},
  journal={Acta Acustica united with Acustica},
  volume={91},
  pages={356--364},
  year={2005}
}

@inproceedings{peebles2023scalable,
  title={Scalable diffusion models with transformers},
  author={Peebles, William and Xie, Saining},
  booktitle={Proceedings of the IEEE/CVF international conference on computer vision},
  pages={4195--4205},
  year={2023}
}

@article{alexanderson2023listen,
  title={Listen, denoise, action! audio-driven motion synthesis with diffusion models},
  author={Alexanderson, Simon and Nagy, Rajmund and Beskow, Jonas and Henter, Gustav Eje},
  journal={ACM Transactions on Graphics (TOG)},
  volume={42},
  number={4},
  pages={1--20},
  year={2023},
  publisher={ACM New York, NY, USA}
}

@article{zhang2024semantic,
  title={Semantic gesticulator: Semantics-aware co-speech gesture synthesis},
  author={Zhang, Zeyi and Ao, Tenglong and Zhang, Yuyao and Gao, Qingzhe and Lin, Chuan and Chen, Baoquan and Liu, Libin},
  journal={ACM Transactions on Graphics (TOG)},
  volume={43},
  number={4},
  pages={1--17},
  year={2024},
  publisher={ACM New York, NY, USA}
}

@InProceedings{Zhi_2023_ICCV,
    author    = {Zhi, Yihao and Cun, Xiaodong and Chen, Xuelin and Shen, Xi and Guo, Wen and Huang, Shaoli and Gao, Shenghua},
    title     = {LivelySpeaker: Towards Semantic-Aware Co-Speech Gesture Generation},
    booktitle = {Proceedings of the IEEE/CVF International Conference on Computer Vision (ICCV)},
    month     = {October},
    year      = {2023},
    pages     = {20807-20817}
}

@inproceedings{liu2025semges,
  title={SemGes: Semantics-aware co-speech gesture generation using semantic coherence and relevance learning},
  author={Liu, Lanmiao and Ghaleb, Esam and Ozyurek, Asli and Yumak, Zerrin},
  booktitle={Proceedings of the IEEE/CVF International Conference on Computer Vision},
  pages={13963--13973},
  year={2025}
}

@inproceedings{liu2022beat,
  title={Beat: A large-scale semantic and emotional multi-modal dataset for conversational gestures synthesis},
  author={Liu, Haiyang and Zhu, Zihao and Iwamoto, Naoya and Peng, Yichen and Li, Zhengqing and Zhou, You and Bozkurt, Elif and Zheng, Bo},
  booktitle={European conference on computer vision},
  pages={612--630},
  year={2022},
  organization={Springer}
}

@inproceedings{cheng2025holegest,
  title={HoleGest: Decoupled Diffusion and Motion Priors for Generating Holisticly Expressive Co-Speech Gestures},
  author={Cheng, Yongkang and Huang, Shaoli},
  booktitle={2025 International Conference on 3D Vision (3DV)},
  pages={748--757},
  year={2025},
  organization={IEEE}
}

@article{ghorbani2023zeroeggs,
  title={ZeroEGGS: Zero-shot Example-based Gesture Generation from Speech},
  author={Ghorbani, Saeed and Ferstl, Ylva and Holden, Daniel and Troje, Nikolaus F and Carbonneau, Marc-Andr{\'e}},
  journal={Computer Graphics Forum},
  volume={42},
  number={1},
  pages={206--216},
  year={2023},
  organization={Wiley Online Library}
}

@inproceedings{li2021audio2gestures,
    title={Audio2Gestures: Generating Diverse Gestures from Speech Audio with Conditional Variational Autoencoders},
    author={Li, Jing and Kang, Di and Pei, Wenjie and Zhe, Xuefei and Zhang, Ying and He, Zhenyu and Bao, Linchao},
    booktitle={Proceedings of the IEEE/CVF International Conference on Computer Vision},
    pages={11293--11302},
    year={2021}
}

@inproceedings{qin2023anyteleop,
  title     = {AnyTeleop: A General Vision-Based Dexterous Robot Arm-Hand Teleoperation System},
  author    = {Qin, Yuzhe and Yang, Wei and Huang, Binghao and Van Wyk, Karl and Su, Hao and Wang, Xiaolong and Chao, Yu-Wei and Fox, Dieter},
  booktitle = {Robotics: Science and Systems},
  year      = {2023}
}

@misc{ze2025gmr,
  title={GMR: General Motion Retargeting},
  author= {Yanjie Ze and João Pedro Araújo and Jiajun Wu and C. Karen Liu},
  year= {2025},
  url= {https://github.com/YanjieZe/GMR},
  note= {GitHub repository}
}

@article{zhang2018egogesture,
  title={EgoGesture: A new dataset and benchmark for egocentric hand gesture recognition},
  author={Zhang, Yifan and Cao, Congqi and Cheng, Jian and Lu, Hanqing},
  journal={IEEE Transactions on Multimedia},
  volume={20},
  number={5},
  pages={1038--1050},
  year={2018},
  publisher={IEEE}
}

@inproceedings{cao2017egocentric,
  title={Egocentric gesture recognition using recurrent 3d convolutional neural networks with spatiotemporal transformer modules},
  author={Cao, Congqi and Zhang, Yifan and Wu, Yi and Lu, Hanqing and Cheng, Jian},
  booktitle={Proceedings of the IEEE international conference on computer vision},
  pages={3763--3771},
  year={2017}
}

@article{indefrey2004spatial,
  title={The spatial and temporal signatures of word production components},
  author={Indefrey, Peter and Levelt, Willem JM},
  journal={Cognition},
  volume={92},
  number={1-2},
  pages={101--144},
  year={2004},
  publisher={Elsevier}
}

@article{comanici2025gemini,
  title={Gemini 2.5: Pushing the frontier with advanced reasoning, multimodality, long context, and next generation agentic capabilities},
  author={Comanici, Gheorghe and Bieber, Eric and Schaekermann, Mike and Pasupat, Ice and Sachdeva, Noveen and Dhillon, Inderjit and Blistein, Marcel and Ram, Ori and Zhang, Dan and Rosen, Evan and others},
  journal={arXiv preprint arXiv:2507.06261},
  year={2025}
}

@article{openai2023gpt4,
  title={GPT-4 Technical Report},
  author={OpenAI},
  journal={arXiv preprint arXiv:2303.08774},
  year={2023}
}

@misc{li2021learn,
      title={AI Choreographer: Music Conditioned 3D Dance Generation with AIST++}, 
      author={Ruilong Li and Shan Yang and David A. Ross and Angjoo Kanazawa},
      year={2021},
      booktitle={ICCV}
}

@inproceedings{radford2021learning,
  title={Learning transferable visual models from natural language supervision},
  author={Radford, Alec and Kim, Jong Wook and Hallacy, Chris and Ramesh, Aditya and Goh, Gabriel and Agarwal, Sandhini and Sastry, Girish and Askell, Amanda and Mishkin, Pamela and Clark, Jack and others},
  booktitle={International conference on machine learning},
  pages={8748--8763},
  year={2021},
  organization={PmLR}
}

@misc{seedtts2024,
  title = {Seed-TTS: A Family of High-Quality Versatile Speech Generation Models},
  author = {Anastassiou, Philip and Chen, Jiawei and Chen, Jitong and Chen, Yuzhou and Liu, Zhenyu and Gu, Jialong and Gao, Shuai and Zhang, Zhe and Li, Changli and Yang, Zengqiang and He, Zhiqi and Zhang, Rui and Qi, Yuancheng and Li, Weiran and Chen, Jian and Zhao, Hangting and Yuan, Yi and Chen, Zehua and Zhao, Liyang and Li, Jianjun and Liu, Shulin and Li, Zhiyuan and Chen, Ming and Zhang, Xinyuan and Wu, Qiushi and Xie, Zhiyong and Wang, Zhengtao and Yan, Li and He, Tao and Chen, Kai and Zhang, Zilong and Huai, Baoxing and Zhang, Zhifu and He, Chong and Lv, Yongqi and Liu, Andi and Wang, Wen and Chen, Yuxuan and Zhang, Xiang and Leng, Yuhang and Zhou, Kai and Zhu, Yuxiang and Li, Huaming and Zou, Chengyi and Wang, Yingjie and Wang, Wenchao and Xiao, Bin and Wang, Tao and Wu, Zhizheng and Qin, Yan},
  year = {2024},
  howpublished = {\url{https://arxiv.org/abs/2406.02430}},
  note = {arXiv:2406.02430 [eess.AS]}
}

@inproceedings{zhang2025semtalk,
  title={SemTalk: Holistic Co-speech Motion Generation with Frame-level Semantic Emphasis},
  author={Zhang, Xiangyue and Li, Jianfang and Zhang, Jiaxu and Dang, Ziqiang and Ren, Jianqiang and Bo, Liefeng and Tu, Zhigang},
  booktitle={Proceedings of the IEEE/CVF International Conference on Computer Vision},
  pages={13761--13771},
  year={2025}
}

@misc{kyutai2024moshi,
      title={Moshi: a speech-text foundation model for real-time dialogue},
      author={Alexandre D\'efossez and Laurent Mazar\'e and Manu Orsini and
      Am\'elie Royer and Patrick P\'erez and Herv\'e J\'egou and Edouard Grave and Neil Zeghidour},
      year={2024},
      eprint={2410.00037},
      archivePrefix={arXiv},
      primaryClass={eess.AS},
      url={https://arxiv.org/abs/2410.00037},
}

@inproceedings{perez2018film,
  title={FiLM: Visual Reasoning with a General Conditioning Layer},
  author={Perez, Ethan and Strub, Florian and De Vries, Harm and Dumoulin, Vincent and Courville, Aaron},
  booktitle={Proceedings of the AAAI Conference on Artificial Intelligence},
  volume={32},
  year={2018}
}

@misc{lipman2023flowmatchinggenerativemodeling,
      title={Flow Matching for Generative Modeling}, 
      author={Yaron Lipman and Ricky T. Q. Chen and Heli Ben-Hamu and Maximilian Nickel and Matt Le},
      year={2023},
      eprint={2210.02747},
      archivePrefix={arXiv},
      primaryClass={cs.LG},
      url={https://arxiv.org/abs/2210.02747}, 
}

@article{qwen3,
    title={Qwen3 Technical Report}, 
    author={An Yang and Anfeng Li and Baosong Yang and Beichen Zhang and Binyuan Hui and Bo Zheng and Bowen Yu and Chang Gao and Chengen Huang and Chenxu Lv and Chujie Zheng and Dayiheng Liu and Fan Zhou and Fei Huang and Feng Hu and Hao Ge and Haoran Wei and Huan Lin and Jialong Tang and Jian Yang and Jianhong Tu and Jianwei Zhang and Jianxin Yang and Jiaxi Yang and Jing Zhou and Jingren Zhou and Junyang Lin and Kai Dang and Keqin Bao and Kexin Yang and Le Yu and Lianghao Deng and Mei Li and Mingfeng Xue and Mingze Li and Pei Zhang and Peng Wang and Qin Zhu and Rui Men and Ruize Gao and Shixuan Liu and Shuang Luo and Tianhao Li and Tianyi Tang and Wenbiao Yin and Xingzhang Ren and Xinyu Wang and Xinyu Zhang and Xuancheng Ren and Yang Fan and Yang Su and Yichang Zhang and Yinger Zhang and Yu Wan and Yuqiong Liu and Zekun Wang and Zeyu Cui and Zhenru Zhang and Zhipeng Zhou and Zihan Qiu},
    journal = {arXiv preprint arXiv:2505.09388},
    year={2025}
}

% Supplementary material is intentionally part of the same PDF and follows the
% main-paper references, as in the supplied reference project.
\clearpage
\appendix
\section*{Supplementary Material}
\addcontentsline{toc}{section}{Supplementary Material}

\section{Motivation, Task Definition, and Contribution Clarification}
\label{supp:task_clarification}

In our manuscript, we formalize the primary objective as achieving \textbf{real-time, semantically aligned, and safety-guaranteed} co-speech gesture generation for humanoid robots, thereby establishing a robust foundation for \textbf{interactive human-humanoid social engagement}.

To further substantiate the technical necessity and the methodological value of our framework, this supplementary material elucidates the core rationales behind our design through three pivotal pillars: (1) the distinct requirements of physical humanoid interaction in Section~\ref{subsection:A.1}, (2) the intrinsic advantages of robot-space learning in Section~\ref{subsection:A.2}, and (3) a granular structural breakdown of our methodological contributions in Section~\ref{subsection:A.3}.

\subsection{Physical Humanoid Interaction vs. Virtual Avatar Animation}
\label{subsection:A.1}
Human-Robot Interaction (HRI) stands at the forefront of Embodied AI, fundamentally branching into physical HRI (pHRI) and social HRI (sHRI). While pHRI traditionally focuses on contact-based manipulation and functional task execution, sHRI demands that robots engage with humans through social intelligence—comforting individuals in distress, providing intuitive gestural guidance, or maintaining interactive companionship. As a physical embodiment, \textbf{a humanoid robot offers a more immersive and multi-modal interaction experience compared to text-based chatbots or screen-bound virtual avatars.} However, this same embodiment \textbf{introduces significant technical hurdles that our work addresses:} specifically the rigorous requirements for real-time streaming, semantic alignment, high-fidelity hand modeling, and, most critically, kinematic executability and safety during physical deployment.

\noindent{\textbf{$\bullet$ Kinematic Safety and Executability.}} Unlike virtual avatar animations that prioritize visual aesthetics without regard for physical constraints, humanoid deployment must strictly adhere to kinematic safety. In real-world scenarios, self-collisions (e.g., the hand striking the torso) or high-frequency jerk (sudden, violent joint accelerations) are not merely visual artifacts but are potentially catastrophic for the robot's hardware and the surrounding human environment. Our framework ensures that every generated gesture is kinematically feasible and physically safe.

\noindent{\textbf{$\bullet$ Low-Latency Streaming Inference.}} While many prior co-speech gesture works rely on offline generation or heavy post-processing, natural Human-Robot Interaction (HRI) demands streaming responsiveness. Excessive latency—where the robot responds only after a significant delay—breaks the social presence and may lead the user to perceive the agent as unresponsive or "broken." Real-time performance is not an elective feature but a fundamental necessity for seamless social engagement.

\noindent{\textbf{$\bullet$ Temporal Semantic Alignment.}} Humans naturally place high cognitive value on semantic gestures (e.g., deictic or iconic motions). When a robot executes the appropriate semantic action at precisely the right temporal onset, it significantly enhances the user's perception of the robot as a truly intelligent social agent. This sense of "agency" is far more profound in the physical world, where the robot’s tangible presence amplifies the impact of its communicative intent compared to a screen-bound avatar.

\noindent{\textbf{$\bullet$ High-Fidelity Hand Modeling.}} In social interaction, the hands are the primary instruments of non-verbal communication, conveying nuances that far exceed the expressive capacity of coarse \textbf{torso or arm movements}. Therefore, we prioritize high-fidelity hand modeling to capture intricate finger articulations, enabling the robot to perform complex social cues and fine-grained semantic gestures that are essential for meaningful human-centric interaction.

% \noindent{\textbf{kinematic Safety.}}  Virtual avatar animation do not care safety concerns like self-collision and jerk. In real life, we can not bear the robot's hand collide with the body or the robot's arm move in a dangerous jerk.

% \noindent{\textbf{Real-time Streaming.}}  Many previous co-speech gesture gneration works focus on offline generation or post-processing, However, interacting human needs low latnecy. If the robot repsond to the human with a long long time, we may think the robot broken. Real-time is a necessity in HRI.

% \noindent{\textbf{Semantic Alignment.}} Human tends to put emphasize on semantic gestures, if a robot can do proper actions in proper time, we may think it is a intelligent agent. This feeling is more strong in physcial world than virtual world.

% \noindent{\textbf{High-fidelity Hand Modeling}.}  If a robot interact with human, the hand is necessary because the hand convey more than 躯干运动. So we need to model the high-fidelity hand. 

% Virtual avatar animation is not constricted with hardware and safety concerns, so it can give various gestures and poses. However, in real world,

\subsection{Why Robot-space Learning is Essential}
\label{subsection:A.2}

A common paradigm in co-speech gesture generation involves generating motions in human skeleton space (Human-space) and subsequently mapping them to robotic hardware via online retargeting and PD control. However, this decoupled approach introduces significant bottlenecks in physical deployment. We argue that learning directly in robot-space is not a matter of experimental convenience, but a fundamental requirement for achieving safe, low-latency, and high-fidelity humanoid interaction.

\noindent{\textbf{$\bullet$ The Problem Chain of Human-space Paradigms.}} Traditional pipelines suffer from a "risk chain" that compromises real-world deployment. First, the \textit{Avatar-to-Humanoid Gap} arises from fundamental differences in joint limits and link lengths; motions generated for human skeletons often fall into singular configurations or exceed the physical torque limits of robot motors. Second, \textit{Latency Amplification} occurs because online retargeting relies on iterative optimization, adding a heavy computational layer that induces perceptible lag. Finally, since human-space models are oblivious to the robot’s physical volume, the resulting trajectories frequently lead to \textit{self-collisions} and numerical "jerk".

\noindent{\textbf{$\bullet$Benefit of Robot-space Learning.}}
 While our motion data originates from human data, RoboGesture internalizes the retargeting process into an offline preprocessing pipeline. It allows for meticulous, optimization-based refinement of joint trajectories—especially for high-dimensional hand articulations—without the stringent temporal constraints of real-time inference. Consequently, this approach yields several advantages: first, it ensures that the evaluation space is perfectly aligned with the deployment space, meaning the model learns a distribution of motions that are natively "robot-aware." By generating robot-specific joint commands end-to-end, we eliminate the need for computationally expensive inference-time retargeting or inverse kinematics, thereby significantly reducing latency. Second, Robot-space learning effectively bypasses the "risk chain" associated with online mapping, such as self-collisions, joint limit violations. 
 
\noindent{\textbf{$\bullet$ Baseline Choice.}} Since the field of co-speech gesture generation tailored specifically for robot-space is currently uncharted, we benchmark against state-of-the-art human-space models. The objective of this comparison is not merely to compete on virtual animation metrics, but to evaluate which paradigm more effectively approaches the requirements of a physical humanoid target setting. In this context, metrics such as \textit{collision rate} and \textit{executability} are treated as primary objectives rather than secondary constraints. By demonstrating the performance gap, we highlight that \textbf{the shift to robot-space is not a matter of experimental convenience, but a necessary prerequisite for moving beyond animation toward robust, real-world embodied intelligence.}

\subsection{Structural Breakdown of Methodological Contributions}
\label{subsection:A.3}

To further clarify the positioning of our work, we categorize the methodological contributions of RoboGesture into three distinct hierarchical levels:

\noindent{\textbf{$\bullet$ Task-level: Streaming Semantic Co-speech Interaction for Humanoids.}} 
Unlike existing works that focus on offline animation or virtual avatar motion synthesis, our task is specifically defined as real-time, streaming, and semantically-aligned interaction on physical humanoid hardware. This shifts the objective from merely generating "human-like" joints to producing "robot-feasible" actions that respond dynamically to human speech. By centering the task on the unique constraints of humanoid robots, we address the practical requirements of low-latency and semantic consistency in physical environments.

\noindent{\textbf{$\bullet$ System-level: Data-Model-Control Co-design.}} 
A key innovation of RoboGesture is the departure from decoupled pipelines. We propose a co-design framework that bridges data, modeling, and control to solve the fragmentation problem inherent in traditional methods. Specifically, we build a robot-centric dataset pre-optimized for physical feasibility, which is used to train a transformer-based generator that inherently understands robot-space joint distributions. This is coupled with a direct-command execution strategy that bypasses risky online retargeting. This holistic integration ensures that the learned features are naturally compatible with the hardware execution layer.

\noindent{\textbf{$\bullet$ Method-level: Safety-aware and Safe Motion Generation.}} 
At the algorithmic level, we introduce three core modules to tackle long-standing technical challenges. First, we address the "modality eclipse"—where models rely too heavily on kinematic history—through an anti-inertia training strategy that forces the model to attend to live audio cues. Second, we design a semantic-acoustic aligner to ensure that generated gestures match both the rhythmic acoustic features and the high-level semantic intent of the speech. Finally, our safety-aware motion generation embeds physical priors directly into the robot-space generation process, enabling the model to natively produce executable trajectories, thereby minimizing the sim-to-real gap.

% Through this three-tiered approach, \textit{RoboGesture} provides a comprehensive solution that moves the field toward a more robust and deployment-ready paradigm for humanoid social interaction.
\section{Real-world Social Interaction Deployment}
\label{supp:real_world}

\begin{figure*}[t]
    \centering
    \includegraphics[trim={0cm 3cm 1cm 2cm}, clip, width=\textwidth]{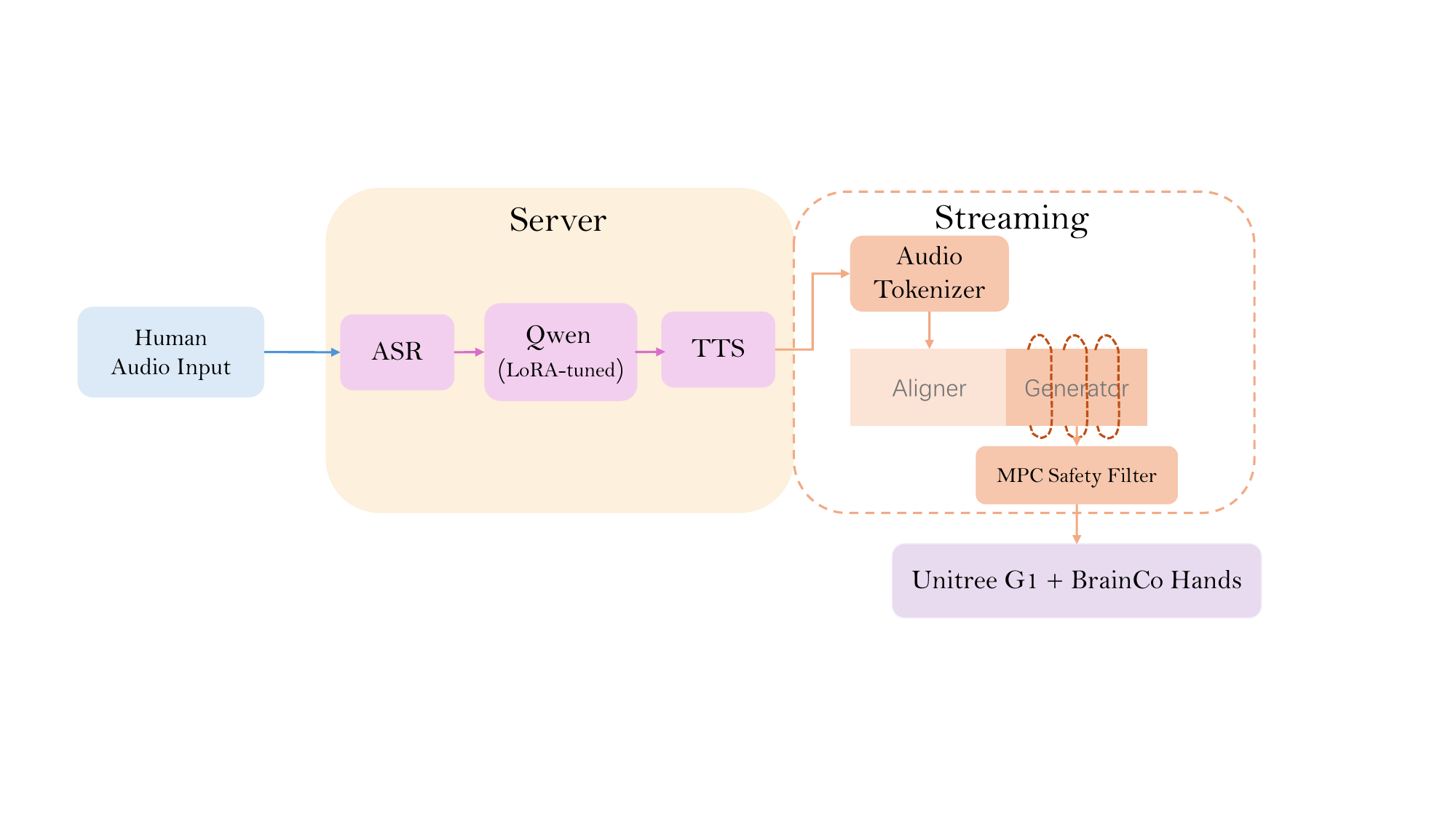}
    \caption{
    Real-world deployment pipeline of our humanoid co-speech interaction system.
    Human speech is first processed by an interaction pipeline including ASR, a LoRA-tuned Qwen module, and TTS.
    The synthesized speech is then fed into the audio tokenizer, semantic-acoustic aligner, and motion generator in a streaming manner.
    The generated motion is further refined by an MPC-based safety filter before being executed on the deployed humanoid platform, i.e., Unitree G1 equipped with BrainCo dexterous hands.
    }
    \label{fig:real_world_inference}
\end{figure*}

This section provides additional details on our real-world deployment pipeline.
Unlike offline gesture-generation systems that only output animation trajectories, our method is designed as an end-to-end interactive system for a physical humanoid platform.
Accordingly, the deployment pipeline must satisfy three practical requirements simultaneously:
(i) \textbf{social responsiveness}, namely producing colloquial and emotionally expressive verbal responses;
(ii) \textbf{streaming motion generation}, namely converting incoming speech into aligned co-speech gestures with low latency;
and (iii) \textbf{physical safety}, namely ensuring that the final motion is executable on the robot without severe self-collision or unstable artifacts.

\subsection{System Overview}

As illustrated in Figure~\ref{fig:real_world_inference}, the deployed system consists of a language interaction module, a streaming speech-to-motion module, and a robot execution module.

\paragraph{Language interaction module.}
Given human speech input, we first perform automatic speech recognition (ASR), then feed the recognized content into a LoRA-tuned Qwen3-4B-Instruct model~\cite{qwen3}, and finally synthesize the textual response into speech through TTS.
This design serves two purposes.
First, it enables open-ended, conversational interaction beyond fixed command-response templates.
Second, and more importantly for our setting, it allows us to generate \emph{socially expressive} responses whose tone and communicative intent are beneficial for downstream co-speech gesture generation.

\paragraph{Streaming speech-to-motion module.}
The synthesized speech and generated motion commands are then processed by the streaming motion stack.
At this stage, the system performs audio tokenization, semantic-acoustic alignment, streaming motion generation, and safety-aware filtering.
The resulting motion commands are then executed on the Unitree G1 humanoid robot equipped with BrainCo dexterous hands.

\subsection{LLM-driven Social Response Generation}

A practical challenge in real-world humanoid interaction is that literal text responses are often insufficient for generating expressive body motion.
Purely factual language tends to produce limited communicative cues, whereas co-speech gestures are more naturally triggered when the response contains clearer social attitude, affective tone, or interpersonal intent.
For this reason, we fine-tune Qwen3-4B-Instruct with LoRA so that it produces conversational responses with explicit affective style tags.

Specifically, the model is trained to prepend each response with an emotion/action tag, such as \texttt{[chat]}, \texttt{[excited]}, or \texttt{[warm]}, followed by the verbal content.
A representative instruction example is shown below:

\begin{quote}
\small
\texttt{
You are an expressive humanoid robot companion. You must respond to the user colloquially with rich emotions. CRITICAL: You MUST begin every response with an emotion/action tag enclosed in brackets (e.g., [chat], [excited], [warm], [angry]), followed by your verbal reply.
}
\end{quote}

For example, given the user query:
\begin{quote}
\small
\texttt{I'm trying to find the small, red book on the top shelf. Do you see it among the others?}
\end{quote}
the fine-tuned model may generate:
\begin{quote}
\small
\texttt{[chat] Let me check. Ah, yes! If you look closely, it's the third one from the left, right next to the blue binder.}
\end{quote}

This design is useful for deployment for two reasons.
First, the emotion/action tag provides an explicit high-level cue for speech style and downstream behavior modulation.
Second, the response style itself becomes more colloquial and socially grounded, which better matches the intended use case of expressive humanoid companionship.
In practice, we feed the generated text into TTS, so that the downstream motion stack receives speech that is not only linguistically meaningful but also socially expressive, thereby improving the natural coupling between language, prosody, and gesture.

\subsection{Runtime Latency Breakdown}

For real-world streaming deployment, the most relevant latency is the time required for each module to produce its \emph{first usable output}, rather than the full completion time of the entire sequence.
Therefore, we report the practical online latency of each key component in Table~\ref{tab:runtime_latency}.
These numbers should be interpreted as module-level streaming latency measurements rather than a strict end-to-end system latency, since practical response time also depends on streaming overlap, chunk availability, and communication overhead.

\begin{table}[t]
\centering
\small
\caption{Runtime latency of the major modules in the deployed system.
For streaming interaction, we focus on the latency to the first usable output.
The motion generator operates on 1-second chunks.}
\label{tab:runtime_latency}
\begin{adjustbox}{max width=\columnwidth}
\begin{tabular}{lc}
\toprule
Module & Online Latency \\
\midrule
ASR & $\max(0.157s, T_{idle})$ \\
Qwen3-4B-Instruct (LoRA-tuned) & 0.142 s \\
TTS & 0.414 s \\
Mimi Audio Tokenizer & 0.161 s \\
Motion Generator & 0.250 s per chunk \\
Filter & 0.170 s per chunk\\
\bottomrule
\end{tabular}
\end{adjustbox}
\end{table}

\paragraph{ASR latency.}
% The ASR module continuously processes the audio stream while the user is speaking.
% When the utterance completes, it requires an average additional latency of $0.157$\,s to finalize the user query, which is then passed to the LLM.
The ASR module processes the audio stream in a streaming fashion. While the computational overhead to finalize an utterance is only $0.157$\,s on average, the actual latency is governed by the endpoint detection threshold $T_{idle}$, which is required to determine if the user has finished speaking. In our implementation, $T_{idle}$ is set to $0.5$\,s. Consequently, the effective latency to finalize a query is $\max(0.157s, T_{idle}) \approx 0.5$\,s, after which the text is passed to the LLM.

\paragraph{LLM latency.}
The LoRA-tuned Qwen3-4B-Instruct model occupies approximately 8430\,MB memory in our setup.
When the model is kept resident in GPU memory (warm start), it achieves an average time-to-first-token (TTFT) of $0.142$\,s, measured over $100$ runs after excluding the initial loading overhead.
This indicates that the language module can provide an interactive response cue with sub-second delay.

\paragraph{TTS latency.}
A streaming TTS module (Seed-TTS~\cite{seedtts2024}) converts the LLM response into the first emotional speech chunk in $0.414$\,s on average.

\paragraph{Streaming audio tokenization latency.}
For the Mimi audio tokenizer~\cite{kyutai2024moshi}, we measure the streaming latency under batch size 1 with the model kept resident in memory.
Over 100 runs after excluding first-load overhead, the average latency to the first streaming token is 0.161\,s.
This is the relevant delay for our pipeline, since it determines how quickly the downstream motion stack can begin processing the generated speech signal.

\paragraph{Motion chunk inference latency.}
The motion generator runs in a chunk-wise streaming manner.
For 1-second motion chunks, we evaluate 240 chunks in total and obtain an average inference latency of 249.78\,ms per chunk.
Equivalently, this corresponds to an effective throughput of 120.11\,fps when counting 30 frames per chunk.
Therefore, once a valid chunk is available, the motion generator itself runs substantially faster than real time.
We note that the 1-second chunk size is an algorithmic context window rather than a fixed additional wall-clock delay; in practice, the time required to obtain the first valid chunk depends on the upstream streaming rate of speech synthesis and tokenization.

\paragraph{MPC-based filter latency.}
Regarding motion control, we implement a streaming MPC-based filter, with detailed formulations provided in Section~\ref{ssec:MPC}. The filter processes the motion stream frame-by-frame with an average execution time of $0.0056$\,s per frame. For a standard $30$\,fps stream, the total processing time for a $1$\,s chunk is approximately $0.1695$\,s, ensuring low-latency, safety-aware gesture execution.

\paragraph{System-level interpretation.}
Taken together, these measurements indicate that the deployed system is practical for streaming humanoid interaction.
The language module and tokenizer provide low-latency first outputs, while the motion generator remains faster than real time under chunk-wise inference.
As a result, the practical interaction delay is primarily governed by the speech pipeline and streaming chunk availability, rather than by the motion model itself.

\subsection{Deployment Safety and Scope}
\label{ssec:deploy_safety}

Our deployment scope is \emph{stationary upper-body co-speech gesturing}: the model outputs arm and dexterous-hand motions only, not locomotion, and lower-body balance is maintained by the built-in Unitree G1 standing controller.
Within this scope, we did not observe instability in practice: across 200 real-world deployment trials, we recorded \textbf{0 falls and 0 emergency stops}.
We note that reinforcement-learning-based whole-body control is important for locomotion-coupled gestures (e.g., gestures that shift the support polygon), and we regard it as a complementary direction for future work rather than a requirement for the upper-body social-gesture setting studied here.

\section{Empirical Study on Modality Eclipse and Anti-Inertia Mechanism}
\label{supp:modality_eclipse}

In the main paper, we describe \emph{modality eclipse} as a practical failure mode in streaming gesture generation: because past motion provides a strong kinematic prior, the generator may over-rely on motion continuity and under-utilize newly arriving semantic-acoustic cues.
Our full framework addresses this issue from multiple aspects, including semantic-acoustic conditioning, Anti-Inertia CFG masking, and an architectural design that delays the dominance of history features in the DiT generator.
This supplementary section provides additional evidence from both a controlled perturbation study and architecture-level ablations.

\begin{table*}[t]
    \centering
    \caption{Ablation study on the key components of our method.
    SA: Semantic Action Score, HD: Hand Detail Score, HN: Human-likeness \& Naturalness, BM: Beat Matching Score.
    Bold indicates the best performance among all configurations.}
    \label{tab:supp_ablation_study}
    \scriptsize
    \begin{tabular*}{0.78\textwidth}{@{\extracolsep{\fill}}l cccc}
        \toprule
        \textbf{Configuration} & SA $\uparrow$ & HD $\uparrow$ & HN $\uparrow$ & BM $\uparrow$ \\
        \midrule
        \rowcolor[gray]{.95} \multicolumn{5}{l}{\textit{Data Scale}} \\
        Ours w/o Semi-data                      & 4.281 & 2.321 & 4.945 & 4.628 \\
        Ours (1/4 Semi-data)                    & 6.316 & 5.980 & 6.882 & 6.892 \\
        \midrule
        \rowcolor[gray]{.95} \multicolumn{5}{l}{\textit{Architecture \& Strategy}} \\
        Ours w/o Context Motion                 & 4.237 & 4.263 & 2.192 & 1.928 \\
        Ours w/o FiLM Injection                 & 5.181 & 4.389 & 4.506 & 4.589 \\
        Ours w/o Semantic Classification        & 5.007 & 4.747 & 4.750 & 5.268 \\
        Ours w/o CFG                            & 4.628 & 4.885 & 6.453 & 6.212 \\
        Ours w/ Reversed History Injection      & 5.376 & 5.136 & 6.094 & 6.100 \\
        AR Strategy                             & 4.843 & 5.031 & 5.358 & 6.850 \\
        \midrule
        \rowcolor[gray]{.95} \multicolumn{5}{l}{\textit{Loss \& Refinement}} \\
        Ours w/o Kinetic-Aware (KA) Loss        & 5.573 & 3.947 & 4.763 & 5.587 \\
        Ours w/o Filter                         & 6.541 & 6.913 & 6.891 & 7.387 \\
        \midrule
        \textbf{Ours (Full Model)}              & \textbf{7.175} & \textbf{7.203} & \textbf{7.394} & \textbf{7.529} \\
        \bottomrule
    \end{tabular*}
    \vspace{-1.0em}
\end{table*}

\subsection{Controlled Perturbation Analysis of History-Dominant Bias}

\paragraph{Diagnostic protocol.}
To directly probe whether the generator is dominated by history or responds to incoming audio, we conduct a controlled perturbation analysis on 50 validation samples.
For each sample, we construct two perturbation settings:
\begin{enumerate}
    \item \textbf{Fixed history, swap audio}: the past-motion context is kept unchanged while the input audio is replaced by another sample.
    \item \textbf{Fixed audio, swap history}: the audio is fixed while the past-motion context is replaced.
\end{enumerate}
We then compute the mean squared error (MSE) between the resulting generated motion sequences.
If swapping history causes a much larger output change than swapping audio, the model exhibits a stronger \emph{history-dominant bias}, indicating shortcut reliance on past motion.

\paragraph{Compared variants.}
We compare the \textbf{Full Model} against an \textbf{Only-DiT} variant that removes the semantic-acoustic conditioning pathway and retains only the generative backbone.
This comparison is intended to isolate whether semantic-acoustic conditioning helps prevent the generator from collapsing into inertial continuation dominated by motion history.

\begin{table*}[t]
\centering
\small
\caption{Controlled perturbation analysis on 50 validation samples.
We report the output MSE when swapping the incoming audio while fixing past motion, or swapping past motion while fixing the audio.
A larger History/Audio ratio indicates a stronger tendency to rely on historical motion rather than incoming audio cues.}
\label{tab:modality_eclipse}
\begin{tabular}{lccc}
\toprule
Model & Audio-swap MSE & History-swap MSE & Hist./Audio Ratio \\
\midrule
Full Model & $0.0619$ & $0.1261$ & $2.04$ \\
Only-DiT   & $0.0312$             & $0.1428$             & $4.57$ \\
\bottomrule
\end{tabular}
\end{table*}

\paragraph{Results.}
Table~\ref{tab:modality_eclipse} shows that both variants are more sensitive to history perturbations than to audio perturbations, confirming that past motion is indeed a dominant control signal in streaming generation.
However, this imbalance is substantially stronger for Only-DiT: its history-to-audio sensitivity ratio reaches $4.57$, compared with $2.04$ for the Full Model.
This indicates that, without semantic-acoustic conditioning, the generator is much more prone to collapse into a history-dominated shortcut, i.e., simply extending prior motion inertia rather than responding to newly arriving audio cues.

Importantly, the Full Model does not eliminate history dependence altogether; instead, it reduces the severity of this bias.
This is consistent with the intended role of our conditioning pathway: motion history remains necessary for temporal continuity, but semantic-acoustic cues provide an additional control source that counterbalances inertial dominance.

\subsection{Audio-Shuffle and Gradient-Magnitude Analysis}
\label{ssec:direct_eclipse}

The perturbation study above probes \emph{output} sensitivity; here we provide two more direct diagnostics on BEAT across three variants---\textbf{Only-DiT}, \textbf{w/o CFG}, and the \textbf{Full} model.
\begin{enumerate}
    \item \textbf{Audio-shuffle FGD.} At inference, we randomly shuffle the input audio against the motion. A model that genuinely relies on audio should degrade sharply (large positive $\Delta$), whereas a model that eclipses audio---or uses it only unstably---changes little.
    \item \textbf{Gradient-magnitude analysis.} We compare $G_{hist}=|\partial\mathcal{L}/\partial E_{hist}|$ against $G_{audio}=|\partial\mathcal{L}/\partial E_{audio}|$. An audio-responsive model should yield a low $G_{hist}/G_{audio}$ ratio with a high absolute $G_{audio}$.
\end{enumerate}

\begin{table*}[t]
\centering
\small
\caption{Direct evidence for modality eclipse on BEAT.
Audio-shuffle FGD measures how much performance degrades when the input audio is shuffled ($\Delta=$ Shuffled $-$ Normal; larger is better).
The gradient analysis reports the sensitivity of the loss to the history vs.\ audio conditions.}
\label{tab:direct_eclipse}
\begin{tabular}{lcccccc}
\toprule
 & \multicolumn{3}{c}{Audio-shuffle FGD} & \multicolumn{3}{c}{Gradient} \\
\cmidrule(lr){2-4}\cmidrule(lr){5-7}
Variant & Normal & Shuffled & $\Delta~\uparrow$ & $G_{hist}$ & $G_{audio}$ & $G_{hist}/G_{audio}~\downarrow$ \\
\midrule
Only-DiT & 1.325 & 1.440 & $+0.11$          & 0.365 & 0.075 & 4.88 \\
w/o CFG  & 1.504 & 1.170 & $-0.33$          & 0.042 & 0.074 & 0.57 \\
Full     & 0.624 & 9.151 & $\mathbf{+8.53}$ & 0.134 & 0.245 & \textbf{0.55} \\
\bottomrule
\end{tabular}
\end{table*}

Table~\ref{tab:direct_eclipse} supports our claim from both angles.
Under audio shuffling, the Full model's FGD jumps from $0.624$ to $9.151$ ($\Delta=+8.53$), confirming that its output is genuinely driven by the audio; in contrast, Only-DiT barely reacts ($\Delta=+0.11$) and \emph{w/o CFG} even improves slightly ($\Delta=-0.33$), the signature of a model that ignores or only unstably uses audio.
The gradient analysis tells the same story: the Full model attains the lowest $G_{hist}/G_{audio}$ ratio ($0.55$) together with the highest absolute audio gradient ($G_{audio}=0.245$), whereas Only-DiT is dominated by history ($G_{hist}/G_{audio}=4.88$).

\paragraph{Drift and the kinetic-aware loss.}
These diagnostics also clarify two design choices.
First, our $15\%$ past-motion masking is precisely an explicit history perturbation: coupled with Anti-Inertia CFG, it drives the gradient ratio from $4.88$ (Only-DiT) down to $0.55$ (Full), grounding rollouts in audio/semantic cues and mitigating drift.
Second, our kinetic-consistency loss is applied to the predicted clean motion $\hat{x}_1 = x_t + (1-t)\,v_\theta$ rather than to the raw noise $x_0$; gating it to $t \ge 0.3$ (i.e., applying it only at low-to-moderate noise) changes test FGD by less than $1\%$, confirming that it is not a source of over-smoothing.

\subsection{Delayed History Injection}

Beyond training-time masking, our architecture also incorporates a simple yet important anti-eclipse design: \emph{Delayed History Injection}.
Instead of allowing past-motion features to dominate the generator from the very beginning, we inject motion-history conditioning only in the later DiT blocks, while the earlier blocks are primarily driven by audio and semantic cues.
Intuitively, this encourages the generator to first establish a response to the current acoustic-semantic input, and only then refine the motion using historical context for temporal continuity.

Concretely, in our DiT generator, the early blocks mainly attend to the audio-conditioned features, while history-conditioned features are introduced in the later stage.
We additionally inject first-layer audio features into the initial block to strengthen the influence of newly arriving audio cues at the beginning of generation.
This design is intended to prevent past motion from prematurely overwhelming the generator, which would otherwise encourage inertial continuation.

To verify that the injection order matters, we perform an additional ablation named \textbf{Ours w/ Reversed History Injection}, where the order is inverted so that past-motion conditioning dominates earlier while audio-guided refinement is pushed later.
As shown in Table~\ref{tab:supp_ablation_study}, reversing the injection order degrades all evaluation metrics, with especially clear drops in semantic action accuracy ($7.175 \rightarrow 5.376$), hand detail ($7.203 \rightarrow 5.136$), and beat matching ($7.529 \rightarrow 6.100$).
This observation supports our hypothesis that premature dominance of motion-history features aggravates modality eclipse.

\subsection{Connection to the Main-Paper Ablations}

The perturbation analysis above should be interpreted as a \emph{diagnostic study} of conditioning bias rather than a standalone quality evaluation.
Its conclusions are complementary to the ablation study in Table~\ref{tab:supp_ablation_study}.
Several observations are particularly relevant.

First, removing semantic-acoustic components consistently harms generation quality:
\emph{w/o FiLM Injection}, \emph{w/o Semantic Classification}, and \emph{Only-DiT} all show substantial drops in semantic action accuracy, hand detail, and naturalness.
Second, removing Anti-Inertia CFG masking (\emph{w/o CFG}) also degrades all four metrics, indicating that training-time mitigation of over-reliance on past motion is beneficial in practice.
Third, reversing the injection order (\emph{w/ Reversed History Injection}) further confirms that not only \emph{what} information is injected, but also \emph{when} it is injected, matters for preventing eclipse by motion history.

Taken together, these results support a coherent picture:
\begin{enumerate}
    \item Streaming gesture generation naturally exhibits a history-dominant bias.
    \item Without semantic-acoustic conditioning, this bias becomes much stronger, as shown by the perturbation study.
    \item Both training-time and architecture-level designs are useful for mitigating this effect: Anti-Inertia CFG masking regularizes shortcut reliance during optimization, while Delayed History Injection prevents past motion from dominating too early inside the generator.
    \item The full design achieves the best overall perceptual and semantic performance because it balances two competing requirements: responsiveness to new semantic-acoustic cues and continuity with historical motion.
\end{enumerate}

We emphasize that the perturbation metric measures \emph{conditioning sensitivity}, rather than whether the model's response to audio is semantically correct in every instance.
For this reason, we use it only as supplementary evidence for the existence of modality eclipse and for the role of semantic-acoustic conditioning in reducing history-dominant shortcut reliance.
The ablation results in Table~\ref{tab:supp_ablation_study} provide the complementary performance-level evidence that the full design not only changes conditioning dynamics, but also improves generation quality in terms of semantic action accuracy, hand expressiveness, naturalness, and beat alignment.

\section{RoboGesture Dataset and Semi-synthetic Data Generation}
\label{supp:robogesture}

% \section{Details on Semi-synthetic Data Generation. }

We briefly described our RoboGesture dataset and semi-synthetic data generation process in Section~4 of the manuscript. In the supplementary material, we provide a detailed explanation of our dataset composition as well as the full data generation pipeline.

\begin{figure*}[t]
  \centering
  \includegraphics[width=0.78\textwidth]{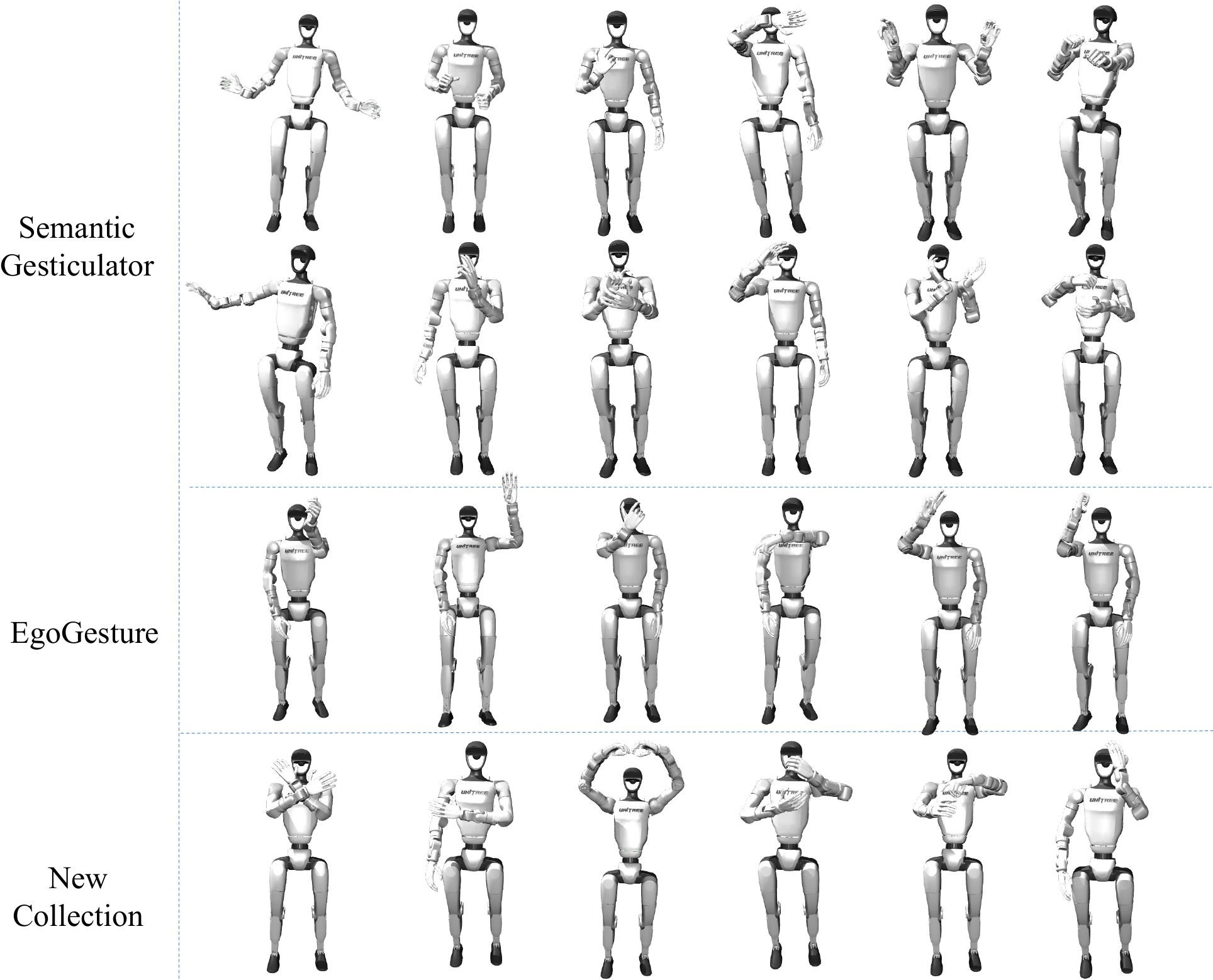}
  \caption{\textbf{Visualization of RoboGesture Dataset.} RoboGesture is a comprehensive full-body gesture dataset comprising over 300 high-quality motion classes collected from EgoGesture, SeG, full-body mocap recordings, online sources, and user surveys. All gestures are retargeted, optimized for physical feasibility, and validated on a real humanoid robot, ensuring human-like expressiveness while remaining fully executable on hardware.}
  \label{fig:supp_robogesture}
\end{figure*}

\subsection{Data Composition}
\label{ssec:data_composition}
In Semantic Gesticulator~\cite{zhang2024semantic}, the SeG dataset provides a rich collection of mocap data, covering a wide range of daily gestures with clear semantics. The downloadable repository includes 544 mocap clips spanning more than 200 semantic gesture classes. We have benefited greatly from this dataset due to its broad coverage and ease of use. However, it also presents several limitations:
(1)Quality Issues. High motion quality is crucial in robotics, where humanoid actions must be safe and collision-free. In SeG, some mocap clips exhibit severe self-collisions—for example, ARMS\_FOLD and FIST\_CLASP —or contain mismatches between the gesture and its semantic label, such as FACE\_COVER and HAND\_FAN.
(2)Embodiment Gap. There remains a significant gap between human motion and what a humanoid robot can perform. Some gestures in SeG rely heavily on head movements that cannot be faithfully reproduced on humanoid robots due to their limited head mobility.
(3)Coverage Limitations. Although SeG includes many daily and culturally relevant gestures, it still lacks coverage of certain gestures required in specific scenarios. To address this gap, we additionally collected new gestures from online sources and user surveys.
To address these limitations, we first filter out motion clips that humanoid robots are fundamentally incapable of performing. We then identify motions with severe physical inconsistencies that cannot be corrected by our MPC-based filters and re-record those clips through motion capture. 

The EgoGesture dataset~\cite{zhang2018egogesture} contains a wide range of hand gestures, defining 83 classes of static and dynamic interactions. It is well-suited for semantic expression tasks due to its rich annotations; however, it focuses solely on hand motion and does not include upper-body or full-body movements. To address this limitation, we re-captured all gesture classes using full-body motion capture, enabling downstream applications that require whole-body expressiveness.

In addition, we collected supplementary gesture data from questionnaires and online sources to cover a broader set of everyday scenarios. These three sources together form our RoboGesture dataset, which contains over 300 high-quality gesture classes, as shown in Figure~\ref{fig:supp_robogesture}.

Each gesture is retargeted to the humanoid model, optimized by our MPC-based filters for physical feasibility, and then replayed on the real robot to verify reachability and control fidelity. The resulting RoboGesture dataset maintains human-like expressiveness while remaining fully executable on the physical robot, providing a robust foundation for downstream gesture synthesis and real-world deployment.

\paragraph{On the gesture taxonomy and data quality.}
The 300+ class list is not defined arbitrarily: it is seeded from the established SeG~\cite{zhang2024semantic} and EgoGesture~\cite{zhang2018egogesture} taxonomies and then expanded with everyday-scenario gestures collected from user surveys, with semantically duplicated or robot-infeasible classes merged or removed during curation.
For quality control, every class is verified at the \emph{motion} level rather than only at the label level: each clip is retargeted, passed through the MPC-based feasibility filter, and replayed on the physical robot, and clips that cannot be made collision-free or reachable are re-recorded via motion capture.
This ensures that the resulting motions are physically plausible and executable at scale, rather than merely correct in label.

\subsection{Human-to-Humanoid Retargeting}
\label{ssec:retarget}

\begin{itemize}
    \item \textbf{Body Motion Retargeting: }We applied the General Motion Retargeting(GMR)~\cite{ze2025gmr} codebase to retarget the human body motion to the robot, with some minor changes. For example, we changed the rotation matrices in the $bvh\_to\_g1.json$ file according to the reset pose of the robot.

    \item \textbf{Hand Motion Retargeting: }We used the dex-retargeting~\cite{qin2023anyteleop} codebase, which maps the fingertip position to the hand joint angle by solving an optimization problem. In our implementation, we did two improvements: \textbf{(1) Non-uniform scaling. } For each finger, we calculate a unique scaling factor, which is the ratio of robot finger length to the human finger length. \textbf{(2) Cartesian Space Alignment. } Dex-retargeting codebase uses the vector from wrist to fingertip. However, the wrist length of the dexterous hand is usually different from that of human hand, resulting in inaccurate mapping. We move the initial point of the vector from the wrist to the MCP point of each finger, thus removing the influence of wrist length.
\end{itemize}

\subsection{MPC-based collision filter}
\label{ssec:MPC}

Given the generated data prepared for our model and the actions predicted by the model, the MPC-based collision filter is applied to prevent self-collision and to smooth the resulting trajectories. We provide the implementation details below.

\begin{itemize}
    \item \textbf{Define the Collison Group:} To save the computation time, we can pick the bodies that require collision check into one collision group. Assuming that there are $M$ bodies in the xml model, and $N$ bodies are selected into one collision pair. In the collison checking process, we only need to do $C_N^2$ distance calculation, while without collision pair there is $C_M^2$ calculation needed. In our implementation, we set three collision groups: \texttt{FINGER\_FINGER\_PAIR}, \texttt{FINGER\_HAND\_PAIR}, \texttt{OTHER\_CHECK\_PAIR}, in order to filter out the collision between fingers, between fingers and hands, and between hands and bodies.
    
    \item \textbf{Set the Collison Avoidance Constraint:} Given two bodies in Mujoco Platform, their minimum distance can be easily obtained. Assume that $dist(q)$ represents the minimum distance between bodies within one collision pair. The collision avoidance constraint can be formulated as:
    \[
    dist(q) \ge dist_{th}
    \]
    where $dist_{th}$ is the collision distance threshold.
    
    \item \textbf{Linearize the Constraint:} We used Taylor expansion to convert the non-linear expression into linear one:
    \[
    dist(q) \approx dist(q_0) + \left. \frac{\partial{dist(q)}}{\partial{q}}\right|_{q=q_0} \cdot\Delta q
    \]
    where $q_0$ is the last angle, $\frac{\partial{dist(q)}}{\partial{q}}$ is the Jacobian matrix, and $\Delta q = q - q_0$. The contact Jacobian matrix can be obtained by the \texttt{mujoco.mj\_jac} API.
    
    \item \textbf{MPC Modeling:} Finally, we formulated this problem as a convex quadratic programming problem:
    % %
    % \[
    % \min_{\Delta q} J(\Delta q) 
    % \]
    % %
    \begin{align*}
    \min_{\Delta q} \quad & J(\Delta q) \\
    \text{subject to} \quad
        & \left.\frac{\partial dist(q)}{\partial q}\right|_{q=q_0}\!\cdot\Delta q \\
        & \hspace{2.8em}\ge dist_{th}+dist(q_0) \\
        & J(\Delta q)=J_1(\Delta q)+J_2(\Delta q) \\
        & \hspace{2.8em}+J_3(\Delta q)
    \end{align*}
    where $J_1(\Delta q)$ denotes the velocity cost, $J_2(\Delta q)$ denotes the trajectory tracking cost, and $J_3(\Delta q)$ denotes the smoothing cost:
    \begin{align*}
        J_1(\Delta q) &= W_1\lVert\Delta q\rVert_2^2, \\
        J_2(\Delta q) &= W_2\lVert q_0+\Delta q-q_{ref}\rVert_2^2, \\
        J_3(\Delta q) &= W_3\lVert q_0+\Delta q-q_{prev}\rVert_2^2.
    \end{align*}

    OSQP Python binding is applied to solve this problem.
\end{itemize}

Notably, our MPC-based collision filter operates on a per-frame basis in a rapid way, inherently supporting streaming processing. This design ensures its utility not only for offline data generation (to curate collision-free training datasets) but also for real-time inference.

\begin{table}[t]
\centering
\small
\caption{Upper-body collision statistics before and after applying the MPC-based safety filter.
We report the percentage of frames containing self-collisions in the upper-body kinematic chain over the generated motion sequences.
The safety filter significantly reduces collision occurrences during real-robot execution.}
\label{tab:collision_filter}
\begin{adjustbox}{max width=\columnwidth}
\begin{tabular}{lc}
\toprule
Method & Collision Frame Ratio (\%) \\
\midrule
Ours (w/o Safety Filter) & 4.16 \\
Ours (with Safety Filter) & \textbf{0.13} \\
\bottomrule
\end{tabular}
\end{adjustbox}
\end{table}

% \subsection{Data Generation Pipeline}

\section{Details on Evaluation Metrics}
\label{sec:details_metrics}

In this section, we provide additional details on the evaluation protocols used in the main paper, including both quantitative metrics and human studies.
Compared with our earlier internal versions, the ECCV submission uses a revised evaluation suite: for quantitative evaluation, we report \textbf{FGD}, \textbf{BC}, \textbf{DIV}, \textbf{MSE}, and \textbf{Col}; for human evaluation, we report \textbf{Rhythm Alignment}, \textbf{Semantic Accuracy}, \textbf{Physical Accuracy \& Hand Consistency}, and \textbf{Overall Preference}.

\subsection{Details on Quantitative Evaluation}
\label{ssec:quant_eval}

\subsubsection{Fréchet Gesture Distance (FGD)}

\textbf{FGD} (Fr\'echet Gesture Distance) evaluates the distributional fidelity between generated motions and reference motions.
It is computed in the latent space of a pre-trained motion autoencoder, following the standard Fr\'echet distance formulation:
\[
\begin{aligned}
\operatorname{FGD}(g,\hat{g})
={}& \lVert\mu_r-\mu_g\rVert_2^2 \\
 &+\operatorname{Tr}\!\Big(\Sigma_r+\Sigma_g
    -2(\Sigma_r\Sigma_g)^{1/2}\Big),
\end{aligned}
\]
where $(g,\hat{g})$ denote real and generated gesture sequences, respectively, and $(\mu_r,\Sigma_r)$ and $(\mu_g,\Sigma_g)$ are the mean and covariance of the latent features extracted from the real and generated motion sets.

\paragraph{Implementation Details.}
We use a dedicated temporal autoencoder as the feature extractor.
The input motion is represented in the 60-DoF robot joint space.
Following the protocol in the main paper, all baseline outputs are first retargeted to our robot morphology before evaluation.
Consistent with our upper-body-focused setting, the first 19 dimensions are excluded during evaluation, and only the remaining 41 dimensions are used to characterize the upper-body gesture quality.
The encoder features are extracted from temporally aligned motion clips, and FGD is computed from the resulting latent distributions.

\subsubsection{Beat Consistency (BC)}

\textbf{BC} (Beat Consistency) measures the temporal synchronization between generated motion and input speech.
Audio beats $B^a$ are extracted from speech onsets, and motion beats $B^m$ are detected as local minima of motion velocity.
The score is defined as:
\[
\mathrm{BC} = \frac{1}{|B^{m}|}\sum_{b_{i}^{m} \in B^{m}}
\exp\!\left(-\frac{\min_{b_{j}^{a} \in B^{a}}|b_{i}^{m}-b_{j}^{a}|^{2}}{2\sigma^{2}}\right),
\]
where $\sigma$ controls the temporal tolerance.

\paragraph{Implementation Details.}
\begin{itemize}
    \item \textbf{Audio Beat Extraction.} We use the onset detection function from Librosa to detect speech onsets, which serve as audio beats $B^a$.
    \item \textbf{Motion Beat Extraction.} Motion beats are computed from the velocity of the 41-dimensional upper-body subset, obtained after excluding the first 19 dimensions from the 60-DoF robot representation. We first compute the frame-wise velocity norm, and then identify local minima as motion beat candidates.
    \item \textbf{Tolerance Parameter.} We set $\sigma = 0.1$ seconds for all methods.
\end{itemize}

\subsubsection{Diversity (DIV)}

\textbf{DIV} measures the diversity of generated motions by computing the average pairwise distance among generated motion clips:
\[
\mathrm{DIV} =
\frac{1}{\binom{N}{2}}
\sum_{i=1}^{N}\sum_{j=i+1}^{N}
\left(
\frac{1}{T \times D}
\sum_{t=1}^{T}\sum_{d=1}^{D}
|p^{i}_{t,d} - p^{j}_{t,d}|
\right),
\]
where $N$ is the number of sampled clips, $T$ is the aligned frame length, and $D$ is the motion dimension.

\paragraph{Implementation Details.}
\begin{itemize}
    \item \textbf{Representation.} DIV is computed directly in the 60-DoF robot joint space.
    \item \textbf{Sampling.} We randomly sample a fixed number of generated clips from the full test set for efficiency.
    \item \textbf{Temporal Alignment.} All sampled clips are aligned to a fixed temporal length before pairwise comparison; shorter clips are zero-padded and longer clips are truncated.
    \item \textbf{Distance.} We use the mean absolute error (MAE) over the full $T \times D$ matrix for each pair.
\end{itemize}

\subsubsection{Mean Squared Error (MSE)}

\textbf{MSE} measures the frame-wise reconstruction error between generated motion and the reference motion:
\[
\mathrm{MSE}(p,\hat{p}) =
\frac{1}{T \times D}
\sum_{t=1}^{T}\sum_{d=1}^{D}
(p_{t,d}-\hat{p}_{t,d})^{2},
\]
where $p$ denotes the reference motion and $\hat{p}$ denotes the generated motion.

\paragraph{Implementation Details.}
\begin{itemize}
    \item \textbf{Evaluation Space.} MSE is computed in the robot joint space after retargeting all compared methods to the same robot morphology.
    \item \textbf{Upper-body Focus.} Consistent with the other motion-quality metrics, we exclude the first 19 dimensions and compute MSE on the remaining 41 upper-body dimensions.
    \item \textbf{Temporal Alignment.} The generated motion is temporally aligned with the reference sequence before evaluation.
    \item \textbf{Scope.} MSE is only reported on datasets where paired reference motion is available; for datasets without motion ground truth, only distributional and human evaluation metrics are reported.
\end{itemize}

\subsubsection{Collision Rate (Col)}

\textbf{Col} measures the physical plausibility of generated motion by quantifying the proportion of frames with self-collision.
Unlike purely kinematic metrics, Col explicitly reflects whether the generated motion is safe for robot execution.

\paragraph{Implementation Details.}
We compute Col using the collision-enabled robot model provided by the GMR/MuJoCo simulation environment.
For each generated motion sequence, we replay the joint trajectory frame by frame in MuJoCo and detect self-contacts using the simulator's built-in collision engine.

\begin{itemize}
    \item \textbf{Collision Geometry.} We use the collision-enabled robot XML corresponding to our humanoid platform, which contains simplified collision primitives for physical contact detection.
    \item \textbf{Upper-body Self-collision Only.} Since the paper focuses on co-speech upper-body gesture generation, we only count collisions between upper-body parts. Contacts involving lower-body bodies (e.g., thigh, calf, ankle, foot) are excluded.
    \item \textbf{Collision Criterion.} For each contact reported by MuJoCo, we treat it as a collision if the contact distance satisfies $\texttt{dist} < \tau$, where $\tau$ is a small negative threshold (set to $-0.01$ in our implementation) to suppress marginal numerical contacts.
    \item \textbf{Frame-level Statistic.} A frame is marked as collided if at least one valid upper-body self-contact is detected.
\end{itemize}

The final collision rate is computed as:
\[
\mathrm{Col} = \frac{N_{\mathrm{collided\_frames}}}{N_{\mathrm{total\_frames}}},
\]
where $N_{\mathrm{collided\_frames}}$ is the number of frames with at least one valid upper-body self-collision and $N_{\mathrm{total\_frames}}$ is the total number of evaluated frames.

\paragraph{Interpretation.}
A lower Col score indicates better physical plausibility and safer deployment behavior.
This metric is particularly important for our setting, since motions that appear visually plausible in animation space may still contain severe upper-body penetrations or unsafe hand--torso interactions after retargeting to a physical humanoid robot.

\subsection{Details on Human Evaluation}
\label{ssec:human_eval_details}

In addition to automatic metrics, we conduct human evaluation to assess perceptual and communicative qualities that cannot be fully captured by distributional or geometric metrics.

\subsubsection{Pairwise Human Evaluation for Main Comparisons}

For the main comparison with baselines, we adopt a pairwise two-alternative forced choice (2AFC) protocol.
We randomly select 20 audio clips from each benchmark, and evaluate our method against the compared baselines under identical audio inputs.
With 5 candidate methods in total (our method plus 4 baselines), each benchmark yields $\binom{5}{2} \times 20 = 200$ pairwise comparison items.

\paragraph{Participants.}
We recruit 100 participants.
To reduce fatigue, each participant evaluates 40 randomly assigned pairwise items.
Under this design, each pair receives approximately 20 independent ratings on average.

\paragraph{Human Metrics.}
Participants are asked to compare the two videos under the following four criteria:
\begin{itemize}
    \item \textbf{Rhythm Alignment}: whether the motion accents, starts, and pauses better match the prosody of the speech.
    \item \textbf{Semantic Accuracy}: whether the gesture content better matches the semantic meaning of the speech.
    \item \textbf{Physical Accuracy \& Hand Consistency}: whether the motion appears physically plausible, free of obvious penetration/jitter, and whether the hand/finger articulation is coherent with the upper-body movement.
    \item \textbf{Overall Preference}: the participant's overall subjective preference.
\end{itemize}

\paragraph{Score Aggregation.}
For each criterion, we convert binary pairwise comparisons into a centered merit score.
Let $V_i$ be the number of pairwise wins for method $i$, and $N_i$ be the number of times it appears in comparisons.
We define:
\[
m_i = 2\frac{V_i}{N_i} - 1,
\]
so that $m_i \in [-1,1]$, where positive values indicate better-than-average preference and negative values indicate worse-than-average preference.
We report the mean merit score together with its standard error across the collected ratings.

\paragraph{Full Results with Uncertainty.}
The complete human-study results used in the main paper are listed below.

% 请确保在文档导言区添加以下两个宏包：
% \usepackage{booktabs}
% \usepackage{graphicx}

\begin{table}[htbp]
    \centering
    % \footnotesize 可以将表格字体调小，如果还需要更小可以使用 \scriptsize
    \footnotesize 
    \caption{Results on the BEAT Benchmark}
    \label{tab:beat_benchmark}
    % 使用 \resizebox 强制表格宽度适应当前行宽（或列宽），高度等比例缩放
    \resizebox{\linewidth}{!}{
        \begin{tabular}{lcccc}
            \toprule
            \textbf{Method} & \textbf{Rhythm Align.} & \textbf{Semantic Acc.} & \textbf{Physical Acc. \& Cons.} & \textbf{Overall Pref.} \\
            \midrule
            Ours                 & $0.1334 \pm 0.0624$  & $0.4270 \pm 0.0702$  & $0.4338 \pm 0.0693$  & $0.3050 \pm 0.0673$ \\
            SemanticGesticulator & $0.0819 \pm 0.1693$  & $0.2264 \pm 0.1834$  & $-0.0273 \pm 0.1708$ & $0.1137 \pm 0.1751$ \\
            SemTalk              & $0.0128 \pm 0.1659$  & $0.0151 \pm 0.1741$  & $0.1679 \pm 0.1827$  & $-0.0132 \pm 0.1624$ \\
            DiffSHEG             & $-0.0058 \pm 0.1526$ & $-0.2691 \pm 0.1972$ & $-0.1440 \pm 0.1943$ & $-0.0992 \pm 0.1885$ \\
            LivelySpeaker        & $-0.2223 \pm 0.1632$ & $-0.3994 \pm 0.1846$ & $-0.4304 \pm 0.1836$ & $-0.3064 \pm 0.1737$ \\
            \bottomrule
        \end{tabular}
    }
\end{table}

\begin{table}[htbp]
    \centering
    \footnotesize
    \caption{Results on the SemanticBEAT Benchmark}
    \label{tab:semanticbeat_benchmark}
    \resizebox{\linewidth}{!}{
        \begin{tabular}{lcccc}
            \toprule
            \textbf{Method} & \textbf{Rhythm Align.} & \textbf{Semantic Acc.} & \textbf{Physical Acc. \& Cons.} & \textbf{Overall Pref.} \\
            \midrule
            Ours                 & $0.1269 \pm 0.0875$  & $0.2239 \pm 0.0880$  & $0.3066 \pm 0.0917$  & $0.2514 \pm 0.0913$ \\
            SemanticGesticulator & $0.0296 \pm 0.0869$  & $0.1605 \pm 0.0873$  & $-0.0347 \pm 0.0922$ & $0.0640 \pm 0.0360$ \\
            SemTalk              & $0.0650 \pm 0.0348$  & $-0.0012 \pm 0.0348$ & $0.1207 \pm 0.0903$  & $0.0970 \pm 0.0901$ \\
            DiffSHEG             & $-0.0090 \pm 0.0888$ & $-0.0817 \pm 0.0890$ & $-0.0458 \pm 0.0360$ & $-0.0626 \pm 0.0922$ \\
            LivelySpeaker        & $-0.2126 \pm 0.0878$ & $-0.3015 \pm 0.0883$ & $-0.3466 \pm 0.0913$ & $-0.3497 \pm 0.0915$ \\
            \bottomrule
        \end{tabular}
    }
\end{table}

\subsubsection{Human Evaluation for Ablation Study}

For the ablation study, the goal is different from the baseline comparison.
Here we aim to evaluate fine-grained design choices under the \emph{same input audio} and within a controlled set of model variants.
Instead of pairwise 2AFC, we therefore adopt a direct scoring protocol.

\paragraph{Protocol.}
For each evaluation batch, participants watch all ablation variants generated from the same audio clip and assign a score from 0 to 10 for each dimension.
To reduce inter-rater scale bias, we further normalize the scores \emph{within each batch}: the top-ranked video is assigned 9 points, the second-ranked video 8 points, and so on.
The normalized scores are then averaged across participants and clips to obtain the final ablation-study scores reported in the main paper.

\paragraph{Ablation Metrics.}
The ablation study uses the same four human dimensions as the main paper:
\begin{itemize}
    \item \textbf{SA}: Semantic Action Score, evaluating whether the performed gesture matches the intended semantic content.
    \item \textbf{HD}: Hand Detail Score, evaluating the richness and clarity of hand/finger articulation.
    \item \textbf{HN}: Human-likeness \& Naturalness, evaluating motion fluency, continuity, and human-like expressiveness.
    \item \textbf{BM}: Beat Matching Score, evaluating the temporal synchronization between gesture rhythm and speech prosody.
\end{itemize}

\paragraph{Rationale.}
This batch-wise ranking protocol is more suitable for ablation analysis than independent absolute scoring, because many ablation variants differ only subtly.
The within-batch normalization suppresses differences in personal score scales and makes the final averaged scores more comparable across participants.

\subsection{Robot-space Baseline Re-training}
\label{ssec:retarget_confound}

Our baselines were originally designed for the human skeleton, so comparing them on our setting requires retargeting their outputs to the robot.
A natural concern is that this retargeting step -- rather than the model itself -- may account for the observed performance gap.
To isolate this factor, we re-train two representative baselines, DiffSHEG~\cite{chen2024diffsheg} and Semantic Gesticulator (SG)~\cite{zhang2024semantic}, \emph{directly in the robot joint space} (i.e., on our robot-space data, with no inference-time retargeting), and compare them against their retargeted counterparts under identical evaluation.
Here Sem.\,BC and Sem.\,DIV denote BC and DIV measured on SemanticBEAT.

\begin{table*}[t]
\centering
\small
\caption{Retargeting is not a confounder.
For each baseline we compare the retargeted variant against a variant re-trained directly in robot joint space.
The two setups yield nearly identical scores, and both trail RoboGesture, indicating that our gains are not attributable to retargeting artifacts.}
\label{tab:retarget_confound}
\begin{tabular}{lccccc}
\toprule
Method & FGD~$\downarrow$ & BC~$\uparrow$ & DIV~$\uparrow$ & Sem.\,BC~$\uparrow$ & Sem.\,DIV~$\uparrow$ \\
\midrule
DiffSHEG (retargeted)     & 2.232 & 0.185 & 0.122 & 0.286 & 0.121 \\
DiffSHEG (direct-trained) & 2.195 & 0.179 & 0.164 & 0.290 & 0.167 \\
SG (retargeted)           & 3.015 & 0.177 & 0.282 & 0.291 & 0.283 \\
SG (direct-trained)       & 2.781 & 0.165 & 0.258 & 0.291 & 0.259 \\
\midrule
RoboGesture (robot-space) & \textbf{0.845} & \textbf{0.187} & 0.208 & \textbf{0.295} & 0.204 \\
\bottomrule
\end{tabular}
\end{table*}

As shown in Table~\ref{tab:retarget_confound}, the directly-trained and retargeted variants of each baseline produce nearly identical scores across all metrics (e.g., FGD $2.195$ vs.\ $2.232$ for DiffSHEG; $2.781$ vs.\ $3.015$ for SG), and both remain far behind RoboGesture (FGD $0.845$).
This confirms that the retargeting step does not materially affect the comparison, and that our improvements stem from the proposed robot-centric framework rather than from retargeting fidelity.

\subsection{Semantic Head Accuracy}
\label{ssec:semantic_acc}

We expect to figure out how discriminative the audio features used by our hierarchical semantic-acoustic aligner are.
We first clarify that Mimi~\cite{kyutai2024moshi} is not a purely acoustic codec: Moshi distills WavLM representations into Mimi's first (semantic) quantizer, which improves phonetic/linguistic discriminability while preserving streaming audio tokens.
On the 300-class gesture-semantic classification task, the semantic head of our aligner reaches \textbf{88.72\%}/\textbf{64.71\%} train/test top-1 accuracy.
Importantly, this 300-class head is only \emph{auxiliary supervision}; the generator finally operates in a continuous DiT latent rather than a 300-way discrete action space.
Because many gesture classes are semantically interchangeable (e.g., different ``greeting'' variants), we additionally report a semantic-equivalent accuracy that credits semantically equivalent predictions.

\begin{table}[t]
\centering
\small
\caption{Accuracy of the 300-class semantic head of our aligner on the test set.
``Strict'' requires the exact class, while ``Semantic-equivalent'' credits semantically interchangeable classes.}
\label{tab:semantic_acc}
\begin{tabular}{lccc}
\toprule
 & Top-1~$\uparrow$ & Top-5~$\uparrow$ & F1~$\uparrow$ \\
\midrule
Strict              & 64.71 & 87.00 & 70.31 \\
Semantic-equivalent & 72.76 & 95.67 & --    \\
\bottomrule
\end{tabular}
\end{table}

\subsection{Text Conditioning and Diversity Analysis}
\label{ssec:text_diversity}

\paragraph{Is text a better semantic interface?}
A reasonable alternative to our finite semantic head is to condition the generator on open-vocabulary text embeddings.
To test this, we replace our aligner with text encoders (Qwen and T5) under the same two-stage protocol, and on 100 SemanticBEAT videos we count the number of identifiable semantic gestures and judge their contextual reasonableness.

\begin{table}[t]
\centering
\small
\caption{Text-conditioning ablation and diversity analysis on 100 SemanticBEAT videos.
We report the number of identifiable semantic gestures and the ratio judged contextually reasonable.
Qwen/T5 replace our aligner with text encoders; SG and DiffSHEG are baselines.}
\label{tab:text_diversity}
\setlength{\tabcolsep}{3.2pt}
\begin{adjustbox}{max width=\columnwidth}
\begin{tabular}{@{}lccccc@{}}
\toprule
 & Ours & Qwen & T5 & SG & DiffSHEG \\
\midrule
Semantic gestures~$\uparrow$ & \textbf{249}    & 145    & 73     & 94     & 73     \\
Reasonable ratio~$\uparrow$  & \textbf{79.0\%} & 21.4\% & 57.0\% & 42.6\% & 41.1\% \\
\bottomrule
\end{tabular}
\end{adjustbox}
\end{table}

As shown in Table~\ref{tab:text_diversity}, our aligner produces more reasonable semantic gestures than the text-encoder variants, indicating that the finite semantic head does not reduce the generator to a fixed set of actions.

\paragraph{On the lower diversity (DIV).}
Our method does not attain the highest global DIV.
We note, however, that SG attains the highest DIV yet a low reasonable ratio, showing that global DIV can be inflated by semantically mismatched motions.
This reframes our lower-but-appropriate DIV as semantically grounded selectivity rather than mode collapse: conditioned on a fixed semantic intent, our generation still exhibits healthy variation while avoiding contextually inappropriate gestures that would otherwise raise the global DIV.

\section{Limitations and Ethical Considerations}
\label{supp:limit}

We conclude by briefly clarifying the scope of the current system, together with several representative failure cases and deployment-related considerations.
Our goal here is not to weaken the main claims of the paper, but to better contextualize what the current system is designed for and where future extensions may be most valuable.

\subsection{Scope and Limitations}

\paragraph{Interaction scope.}
The present work focuses on \emph{upper-body co-speech gesture generation} for humanoid interaction.
Accordingly, the system is designed for speech-aligned expressive motion, rather than full-body long-horizon embodied behavior such as locomotion, object manipulation, or scene-level decision making.
We view this as an important but still scoped step toward broader humanoid social intelligence.

\paragraph{History-conditioned generation.}
Our perturbation analysis shows that past motion remains an important control signal in streaming generation.
The proposed semantic-acoustic conditioning, Anti-Inertia CFG masking, and delayed history injection reduce excessive history dominance, but do not entirely remove the need for motion continuity priors.
This is expected in practice: good co-speech motion must balance responsiveness to new audio cues with temporal smoothness across consecutive chunks.

\paragraph{Data and generalization.}
A substantial part of our training pipeline relies on robot-space semi-synthetic data construction.
This design is helpful for improving semantic coverage and deployment consistency, but the resulting model is still influenced by the diversity of the constructed data distribution.
Further scaling with richer real-world interactive recordings would likely improve robustness in broader social scenarios.

\paragraph{System latency.}
As discussed in the deployment section, the motion model itself runs faster than real time once chunked audio is available.
In the current prototype, the dominant delay comes from the overall interaction pipeline, including language generation, speech synthesis, and chunked streaming.
Reducing such system-level latency is an important direction for future deployment-oriented work.

\paragraph{Representative examples.}
Representative examples of successful deployment and typical failure cases are included in the supplementary video.

\subsection{Ethical Notes}

\paragraph{Real-world deployment.}
Since the generated motion is executed on a physical humanoid robot, real-world use should always be accompanied by standard safety measures such as conservative control bounds, monitoring, and emergency stop mechanisms.
Our safety filter improves executability, but it should be understood as part of a broader deployment stack rather than a complete substitute for system-level safety practice.

\paragraph{Expressive social interaction.}
The purpose of this work is to improve the naturalness and communicative quality of humanoid interaction.
At the same time, more expressive robot behavior can shape user expectations more strongly than purely functional systems.
We therefore believe that practical deployment should remain transparent about the system's generated nature and intended application scope.

\paragraph{Future outlook.}
We hope this work can serve as a useful step toward expressive and deployable humanoid interaction, while also motivating future research on richer real-world data, lower-latency system integration, and stronger deployment-time safeguards.

\end{document}